\documentclass{article}

\usepackage[sort&compress,numbers]{natbib} 

\usepackage{microtype}
\usepackage{graphicx}
\usepackage{booktabs} 
\usepackage{amsmath} 
\usepackage{amsfonts}
\usepackage{mathtools}
\usepackage{colortbl}
\usepackage{subfig}

\usepackage{hyperref}

\usepackage[accepted]{icml2021}

\definecolor{Blue}{RGB}{3, 31, 97}
\definecolor{Blue1}{RGB}{214, 235, 245}
\definecolor{Blue2}{RGB}{235, 245, 250}
\definecolor{Gray}{RGB}{247, 252, 255}

\begin{document}

\icmltitlerunning{Generative Adversarial Transformers}










\definecolor{darkblue}{RGB}{21,0,133}

\newcommand\blfootnote[1]{%
   \begingroup
   \renewcommand\thefootnote{}\footnote{#1}%
   \addtocounter{footnote}{-1}%
   \endgroup
}

\twocolumn[
\title{Generative Adversarial Transformers}

\author{Drew A. Hudson$^{\S}$ \\
Department of Computer Science\\
Stanford University\\
\texttt{dorarad@cs.stanford.edu}\\
\And
C. Lawrence Zitnick \\
Facebook AI Research\\
Facebook, Inc.\\
\texttt{zitnick@fb.com}\\
}

\icmlkeywords{Machine Learning, ICML, generative models, GANs, transformers, image synthesis, attention, compositionality}

\vspace{-8pt}
\maketitle


]




\printAffiliationsAndNotice{} 


\begin{abstract}
We introduce the GANformer, a novel and efficient type of transformer, and explore it for the task of visual generative modeling. The network employs a bipartite structure that enables long-range interactions across the image, while maintaining computation of linear efficiency, that can readily scale to high-resolution synthesis. It iteratively propagates information from a set of latent variables to the evolving visual features and vice versa, to support the refinement of each in light of the other, and encourage the emergence of compositional representations for objects and scenes. In contrast to the classic transformer architecture, it utilizes multiplicative integration that allows flexible region-based modulation, and can thus be seen as a multi-latent generalization of the successful StyleGAN network. We demonstrate the model's strength and robustness through a careful evaluation over a range of datasets, from simulated multi-object environments to rich real-world indoor and outdoor scenes, showing it attains state-of-the-art results in terms of image quality and diversity, while enjoying fast learning and better data-efficiency. Further qualitative and quantitative experiments offer an insight into the model's inner workings, revealing improved interpretability and stronger disentanglement, and illustrate the benefits and efficacy of our approach. An implementation of the model is available at \url{https://github.com/dorarad/gansformer}.
\end{abstract}

\begin{figure}[t]
\centering
\subfloat{\includegraphics[width=0.3303333\linewidth]{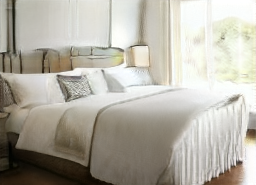}}
\hfill
\subfloat{\includegraphics[width=0.3303333\linewidth]{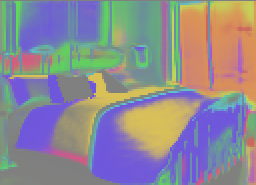}}
\hfill
\subfloat{\includegraphics[width=0.3303333\linewidth]{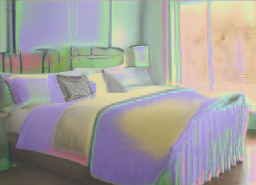}}
\vspace*{-9.5pt}
\subfloat{\includegraphics[width=0.3303333\linewidth]{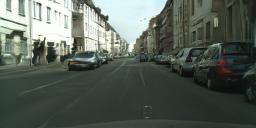}}
\hfill
\subfloat{\includegraphics[width=0.3303333\linewidth]{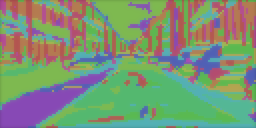}}
\hfill
\subfloat{\includegraphics[width=0.3303333\linewidth]{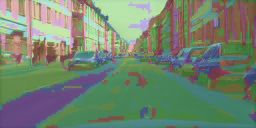}}
\vspace*{-9.5pt}
\subfloat{\includegraphics[width=0.3303333\linewidth]{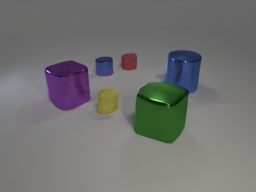}}
\hfill
\subfloat{\includegraphics[width=0.3303333\linewidth]{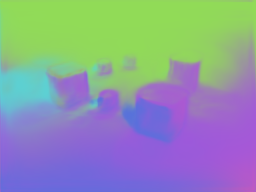}}
\hfill
\subfloat{\includegraphics[width=0.3303333\linewidth]{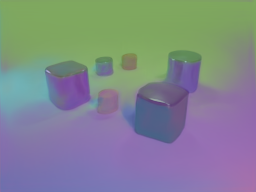}}
\vspace*{-5.5pt}
\caption{Sample images generated by the GANformer, along with a visualization of the model attention maps.}
\vspace*{-17pt}
\label{teaser}
\end{figure} 

\vspace*{-22pt}
\section{Introduction}
\label{intro}
The cognitive science literature speaks of two reciprocal mechanisms that underlie human perception: the \textbf{\textit{bottom-up}} processing, proceeding from the retina up to the cortex, as local elements and salient stimuli hierarchically group together to form the whole \citep{gibson1}, and the \textbf{\textit{top-down}} processing, where surrounding global context, selective attention and prior knowledge inform the interpretation of the particular \citep{gregory1}. While their respective roles and dynamics are being actively studied, researchers agree that it is the interplay between these two complementary processes that enables the formation of our rich internal representations, allowing us to perceive the world around in its fullest and create vivid imageries in our mind's eye \citep{vision1,vision2,vision3,vision4}\blfootnote{$^{\S}${\color{darkblue}I wish to thank Christopher D. Manning for the fruitful discussions and constructive feedback in developing the bipartite transformer, especially when explored within the language representation area, as well as for the kind financial support that allowed this work to happen.}
}. 

\begin{figure*}[ht]
\centering
\subfloat{\includegraphics[width=0.775\linewidth]{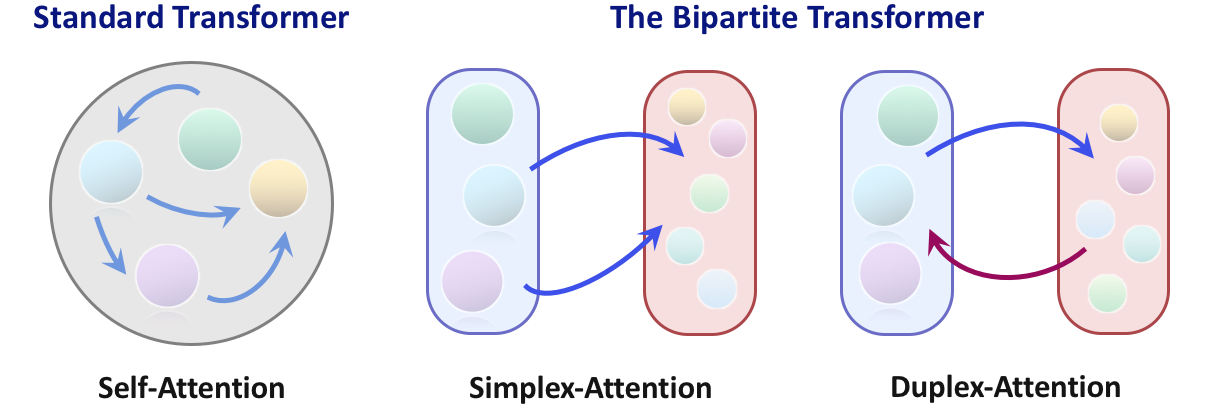}}
\caption{\textbf{Bipartite Attention.} We introduce the GANformer network, that leverages a bipartite structure to support long-range interactions while evading the quadratic complexity standard transformers suffer from. We present two novel attention operations over the bipartite graph: \textit{simplex} and \textit{duplex}, the former permits communication in one direction, in the generative context -- from the latents to the image features, while the latter enables both top-down and bottom-up connections between these two dual representations.}
\label{plots}
\vspace*{-8pt}
\end{figure*}

Nevertheless, the very mainstay and foundation of computer vision over the last decade -- the Convolutional Neural Network, surprisingly, does not reflect this bidirectional nature that so characterizes the human visual system, and rather displays a one-way feed-forward progression from raw sensory signals to higher representations. Unfortunately, the local receptive field and rigid computation of CNNs reduce their ability to model long-range dependencies or develop holistic understanding of global shapes and structures that goes beyond the brittle reliance on texture \citep{texture}, and in the generative domain especially, they are linked to considerable optimization and stability issues \citep{sagan} due to their fundamental difficulty in coordinating between fine details across the generated scene. These concerns, along with the inevitable comparison to cognitive visual processes, beg the question of whether convolution alone provides a complete solution, or some key ingredients are still missing. 

Meanwhile, the NLP community has witnessed a major revolution with the advent of the Transformer network \citep{transformer}, a highly-adaptive architecture centered around relational attention and dynamic interaction. In response, several attempts have been made to integrate the transformer into computer vision models, but so far they have met only limited success due to scalabillity limitations stemming from its quadratic mode of operation.

Motivated to address these shortcomings and unlock the full potential of this promising network for the field of computer vision, we introduce the Generative Adversarial Transformer, or GANformer for short, a simple yet effective generalization of the vanilla transformer, explored here for the task of visual synthesis. The model utilizes a bipartite structure for computing soft attention, that iteratively aggregates and disseminates information between the generated image features and a compact set of latent variables that functions as a \textit{\textbf{bottleneck}}, to enable bidirectional interaction between these dual representations. This design achieves a favorable balance, being capable of flexibly modeling global phenomena and long-range interactions on the one hand, while featuring an efficient setup that still scales linearly with the input size on the other. As such, the GANformer can sidestep the computational costs and applicability constraints incurred by prior works, caused by the dense and potentially excessive pairwise connectivity of the standard transformer \citep{sagan,biggan}, and successfully advance the generative modeling of compositional images and scenes. 

We study the model's quantitative and qualitative behavior through a series of experiments, where it achieves state-of-the-art performance for a wide selection of datasets, of both simulated as well as real-world kinds, obtaining particularly impressive gains in generating highly-structured multi-object scenes. As indicated by our analysis, the GANformer requires less training steps and fewer samples than competing approaches to successfully synthesize images of high quality and diversity. Further evaluation provides robust evidence for the network's enhanced transparency and compositionality, while ablation studies empirically validate the value and effectiveness of our approach. We then present visualizations of the model's produced attention maps, to shed more light upon its internal representations and synthesis process. All in all, as we will see through the rest of the paper, by bringing the renowned GANs and Transformer architectures together under one roof, we can integrate their complementary strengths, to create a strong, compositional and efficient network for visual generative modeling.
\section{Related Work}
\label{related}

Generative Adversarial Networks (GANs) \citep{gan}, originally introduced in 2014, have made remarkable progress over the past  years, with significant advances in training stability and dramatic improvements in image quality and diversity. that turned them to be nowadays one of the leading paradigms in visual synthesis \citep{dcgan,biggan,stylegan}. In turn, GANs have been widely adopted for a rich variety of tasks, including image-to-image translation \citep{pix2pix,cyclegan}, super-resolution \citep{superres}, style transfer \citep{stargan}, and representation learning \citep{bigan}, to name a few. But while generated images for faces, single objects or natural scenery have reached astonishing fidelity, becoming nearly indistinguishable from real samples, the unconditional synthesis of more structured or compositional scenes is still lagging behind, suffering from inferior coherence, reduced geometric consistency and, at times, a lack of global coordination \citep{johnson,iclrscenes,sagan}. As of now, faithful generation of structured scenes is thus yet to be reached. 

Concurrently, the last years saw impressive progress in the field of NLP, driven by the innovative architecture called Transformer \citep{transformer}, which has attained substantial gains within the language domain and consequently sparked considerable interest across the deep learning community \citep{transformer,bert}. In response, several attempts have been made to incorporate self-attention constructions into vision models, most commonly for image recognition, but also in segmentation \citep{attsgm}, detection \citep{detr}, and synthesis \citep{sagan}. From structural perspective, these can be roughly divided into two streams: those that apply local attention operations, failing to capture global interactions \citep{localtrns1,localtrns2,localtrns3,localtrns4,imgtrns}, and others that borrow the original transformer structure as-is and perform attention globally across the entire image, resulting in prohibitive computation due to the quadratic complexity, which fundamentally hinders its applicability to low-resolution layers only \citep{sagan,biggan,glbltrns1,glbltrns2,taming,vistrns,transgan}. Few other works proposed sparse, discrete or approximated variations of self-attention, either within the adversarial or autoregressive contexts, but they still fall short of reducing memory footprint and computational costs to a sufficient degree \citep{gsa,axial,criscros,taming,sparsetrns}. 

Compared to these prior works, the GANformer stands out as it manages to avoid the high costs ensued by {\textit{self attention}, employing instead \textbf{\textit{bipartite attention}} between the image features and a small collection of latent variables. Its design fits naturally with the generative objective of transforming source latents into an output image, facilitating long-range interaction without sacrificing computational efficiency. Rather, the network maintains a scalable linear computation across all layers, realizing the transformer's full potential. In doing so, we seek to take a step forward in tackling the challenging task of scene generation. Intuitively, and as is later corroborated by our findings, allocating multiple latents to interact through attention with the generated image serves as a structural prior of a bottleneck that promotes the formation of compact and compositional scene representations, as the different latents may specialize to certain objects or semantic regions of interest. Indeed, as demonstrated in section \ref{exps}, the Generative Adversarial Transformer achieves state-of-the-art performance in synthesizing varied real-world indoor and outdoor scenes, while showing indications for semantic disentanglement along the way. 

In designing our model, we draw inspiration from multiple lines of research on generative modeling, compositionality and scene understanding, including techniques for scene decomposition, object discovery and representation learning. Several variational approaches \citep{monet,iodine,air,genesis} perform iterative inference to encode scenes into multiple slots, but are mostly applied in the contexts of synthetic and oftentimes fairly rudimentary 2D settings. Works such as Capsule networks \citep{capsules1,rim} leverage ideas from psychology about Gestalt principles \citep{gestalt1,gestalt2}, perceptual grouping \citep{gestalt3} or analysis-by-synthesis \cite{rbc1}, and like us, introduce ways to piece together visual elements to discover compound entities and, in the cases of Set Transformers \citep{settrns} or $A^2$-Nets \citep{doubleatt}, group local information into global aggregators, which proves useful for a broad spectrum of tasks, spanning unsupervised segmentation \citep{nem,slotatt}, clustering \citep{settrns}, image recognition \citep{lambda}, NLP \citep{etc} and viewpoint generalization \citep{capsules3}. However, our work stands out incorporating new ways to integrate information across the network through novel forms of attention: (\textit{Simplex} and \textit{Duplex}), that iteratively update and refine the assignments between image features and latents, and is the first to explore these techniques in the context of high-resolution generative modeling.

\begin{figure}[t]
\centering
\subfloat{\includegraphics[width=1\linewidth]{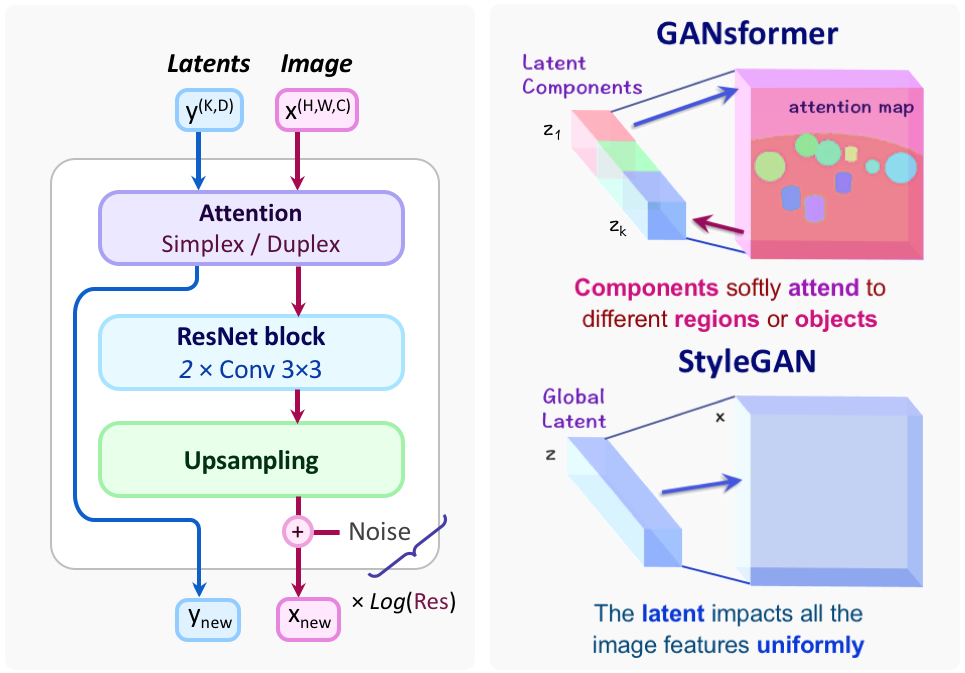}}
\small
\caption{\textbf{Model overview.} {\textbf{Left}}: The GANformer layer is composed of a bipartite attention operation to propagate information from the latents to the image grid, followed by convolution and upsampling. These are stacked multiple times starting from an initial $4{\times}4$ grid and up to producing a final high-resolution image. {\textbf{Right}}: The latents and image features attend to each other to capture the scene structure. The GANformer's compositional latent space contrasts with the StyleGAN's monolithic one (where a single latent modulates the whole scene uniformly).}
\vspace*{-16pt}
\label{modelover}
\end{figure}

\begin{figure*}[t]
\centering
\subfloat{\includegraphics[width=0.14\linewidth]{images/69a.png}}
\hfill
\subfloat{\includegraphics[width=0.14\linewidth]{images/69b.png}}
\hfill
\subfloat{\includegraphics[width=0.14\linewidth]{images/69c.png}}
\hfill
\subfloat{\includegraphics[width=0.193\linewidth]{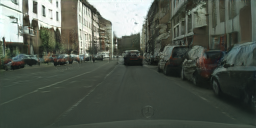}}
\hfill
\subfloat{\includegraphics[width=0.193\linewidth]{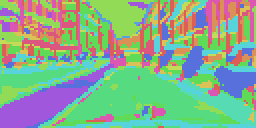}}
\hfill
\subfloat{\includegraphics[width=0.193\linewidth]{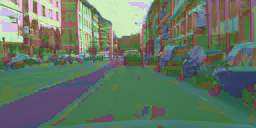}}
\vspace*{-10pt}
\centering
\subfloat{\includegraphics[width=0.166\linewidth]{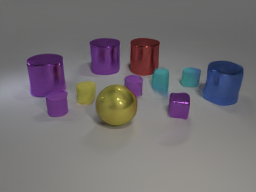}}
\hfill
\subfloat{\includegraphics[width=0.166\linewidth]{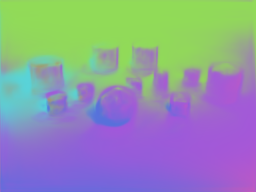}}
\hfill
\subfloat{\includegraphics[width=0.166\linewidth]{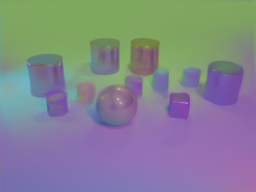}}
\hfill
\subfloat{\includegraphics[width=0.166\linewidth]{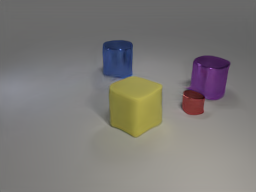}}
\hfill
\subfloat{\includegraphics[width=0.166\linewidth]{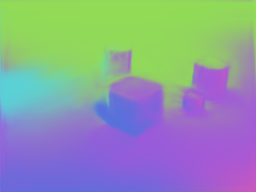}}
\hfill
\subfloat{\includegraphics[width=0.166\linewidth]{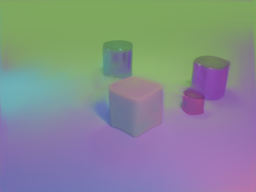}}
\vspace*{-4pt}
\caption{\textbf{Attention maps.} Sample images generated by the GANformer for the CLEVR, LSUN-Bedrooms and Cityscapes datasets, and a visualization of the produced attention maps, from lower (\textbf{top row}) and upper (\textbf{bottom row}) layers. The colors correspond to the different latents that attend to each region.}
\label{imgs}
\vspace*{-8pt}
\end{figure*}

Most related to our work are certain GAN models for conditional and unconditional visual synthesis: A few methods \citep{kgan,mganprior,blockgan,relate} utilize multiple replicas of a generator to produce a set of image layers, that are then combined through alpha-composition. As a result, these models make quite strong assumptions about the independence between the components depicted by each layer. In contrast, our model generates one unified image through a cooperative process, coordinating between the different latents through the use of soft attention. Other works, such as SPADE \citep{spade,sean}, employ region-based feature modulation for the task of layout-to-image translation, but, contrary to us, use fixed segmentation maps and static class embeddings to control the visual features. Of particular relevance is the prominent StyleGAN model \citep{stylegan,stylegan2}, which utilizes a single global style vector to consistently modulate the features of each layer. The GANformer generalizes this design, as multiple style vectors impact different regions in the image concurrently, allowing for spatially finer control over the generation process. Finally, while StyleGAN broadcasts information in one direction from the single \textit{global} latent to the \textit{local} image features, our model propagates information both from latents to features and vice versa, enabling top-down and bottom-up reasoning to occur simultaneously\footnote{Note however that our model certainly does not claim to serve as a biologically-accurate reflection of cognitive top-down processing. Rather, this analogy plays as a conceptual source of inspiration that aided us through the idea development.}.
%
\section{The Generative Adversarial Transformer}
\label{model}

The Generative Adversarial Transformer (\textit{GANformer}) is a type of Generative Adversarial Network, which involves a \textit{generator} network (G) that maps random samples from the \textit{latent} space to the output space (e.g. an image), and a \textit{discriminator} network (D) which seeks to discern between real and fake samples \citep{gan}. The two networks compete with each other through a minimax game until reaching an equilibrium. Typically, each of these networks consists of multiple layers of convolution, but in the GANformer case, we instead construct them using a novel architecture, called 
\textbf{\textit{Bipartite Transformer}}, formally defined below. 

The section is structured as follows: we first present a formulation of the Bipartite Transformer, a domain-agnostic generalization of the Transformer\footnote{By \textit{transformer}, we precisely mean a multi-layer bidirectional transformer encoder, as described in \citep{bert}, which interleaves self-attention and feed-forward layers.}  (\textbf{section \ref{definition}}). Then, we provide an overview of how the transformer is incorporated into the generative adversarial framework (\textbf{section \ref{gan}}). We conclude by discussing the merits and distinctive properties of the GANformer, that set it apart from the traditional GAN and transformer networks (\textbf{section \ref{discussion}}).

\subsection{The Bipartite Transformer}
\label{definition}

\begin{figure*}[ht]
\centering
\subfloat{\includegraphics[width=0.142\linewidth]{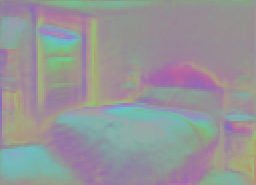}}
\hfill
\subfloat{\includegraphics[width=0.142\linewidth]{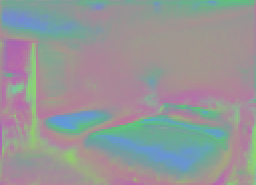}}
\hfill
\subfloat{\includegraphics[width=0.142\linewidth]{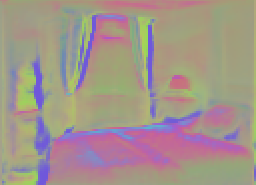}}
\hfill
\subfloat{\includegraphics[width=0.142\linewidth]{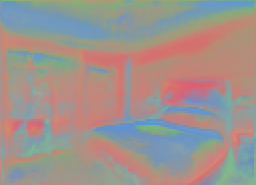}}
\hfill
\subfloat{\includegraphics[width=0.142\linewidth]{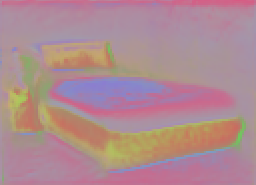}}
\hfill
\subfloat{\includegraphics[width=0.142\linewidth]{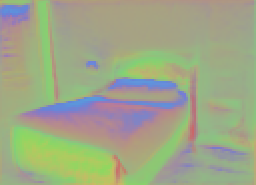}}
\hfill
\subfloat{\includegraphics[width=0.142\linewidth]{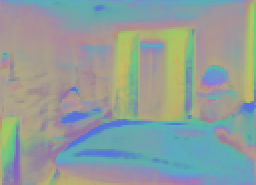}}
\vspace*{-5pt}
\caption{\footnotesize \textbf{Upper-layer attention maps} produced by the GANformer model during synthesis, for the LSUN-Bedrooms dataset.}
\label{attmaps}
\vspace*{-12pt}
\end{figure*}


The standard transformer network is composed of alternating multi-head self-attention and feed-forward layers. We refer to each pair of self-attention and feed-forward operations as a transformer layer, such that a \textit{transformer} is considered to be a stack of several such layers. The Self-Attention layer considers all pairwise relations among the input elements, updating each one by attending to all the others. The Bipartite Transformer generalizes this formulation, featuring instead a bipartite graph between two groups of variables -- in the GAN case, latents and image features. In the following, we consider two forms of attention that could be computed over the bipartite graph -- Simplex attention and Duplex attention, depending on the direction in which information propagates\footnote{In computer networks, \textit{simplex} refers to single direction communication, while \textit{duplex} refers to communication in both ways.} -- either in one way only, from the latents to the image, or both in top-down and bottom-up ways. While for clarity purposes, we present the technique here in its one-head version, in practice we make use of a multi-head variant, in accordance with prior work \citep{transformer}. 

\subsubsection{Simplex Attention}
We begin by introducing the \textbf{\textit{simplex attention}}, which distributes information in a single direction over the bipartite transformer graph. Formally, let ${X}^{n{\times}d}$ denote an input set of $n$ vectors of dimension $d$ (where, for the image case, $n = W{\times}H$), and ${Y}^{m{\times}d}$ denote a set of $m$ aggregator variables (the latents, in the generative case). We can then compute attention over the derived bipartite graph between these two groups of elements. Specifically, we define:
\begin{gather*} 
{Attention}(Q,K,V) = \text{softmax}\left(\frac{QK^{T}}{\sqrt{d}}\right) V \\
{a}(X,Y) = {Attention}(q(X),k(Y),v(Y))
\end{gather*}
Where $q(\cdot),k(\cdot),v(\cdot)$ are functions that respectively map elements into queries, keys, and values, all maintaining dimensionality $d$. We also provide the mappings with positional encodings, to reflect the distinct spatial position of each element e.g. in the image (see section \ref{gan} for details). Note that this \textbf{\textit{bipratite attention}} is a generalization of self attention, where $Y=X$.  

We can then integrate the attended information with the input elements ${X}$, but whereas the standard transformer implements an additive update rule of the form: 
$${u}^{a}(X,Y) = {LayerNorm}(X + {a}({X},{Y}))$$ 
we instead use the retrieved information to control both the \textit{scale} as well as the \textit{bias} of the elements in $X$, in line with the practice promoted by the StyleGAN model \citep{stylegan}. As our experiments indicate, such multiplicative integration enables significant gains in the model's performance. Formally:
$$ {u}^{s}({X},{Y}) = \gamma\left({a}({X},{Y})\right) \odot \omega({X}) + \beta\left({a}({X},{Y})\right) $$
Where $\gamma(\cdot),\beta(\cdot)$ are mappings that compute multiplicative and additive factors (scale and bias), both maintaining a dimension of $d$, and $\omega({X}) = \frac{{X}-\mu({X})}{\sigma({X})}$ normalizes the features of $X$\footnote{The statistics are computed either with respect to other elements in $X$ for instance normalization, or among element channels in the case of layer normalization, which performs better.}. By normalizing $X$ (the image features), and then letting $Y$ (the latents) control $X$'s statistical tendencies, we essentially enable information propagation from $Y$ to $X$, intuitively, allowing the latents to control the visual generation of spatial attended regions within the image, so as to guide the synthesis of objects and entities. 

\subsubsection{Duplex Attention}
We can go further and consider the variables $Y$ to posses a key-value structure of their own \citep{kvmn}: $Y = (K^{m{\times}d},V^{m{\times}d})$, where the \textit{values} V store the \textit{{content}} of the $Y$ variables as before (i.e. the randomly sampled latent vectors) while the \textit{keys} K track the \textbf{\textit{centroids}} of the attention-based assignment between $X$ and $Y$, which can be computed by ${K} = {a}(Y,X)$ -- namely, the weighted averages over $X$ elements, using the attention distribution derived by comparing them to $Y$ elements. Intuitively, each centroid tracks the region in the image $X$ that interacts with the respective latent in $Y$. Consequently, we can define a new update rule:
$$ {u}^{d}({X},{Y}) = \gamma({A}(Q,K,V)) \odot \omega({X}) + \beta({A}(Q,K,V)) $$
This update compounds together two attention operations: first \textbf{\color{darkblue} (1)} computing  attention assignments between $X$ and $Y$, by $K={a}(Y,X)$, and then \textbf{\color{darkblue} (2)} refining the soft assignments by considering their centroids, through ${A}(Q,K,V)$, where $Q=q(X)$, which computes attention between the elements $X$ and their centoroids $K$. This is analogous to the Expectation–Maximization or k-means algorithms, \citep{kmeans,slotatt}, where we iteratively refine the assignments of elements $X$ to clusters $Y$ based on their distance to their respective centroids $K={a}(Y,X)$. As is empirically shown later, this works more effectively than the update ${u}^{s}$ defined above. 

Finally, to support bidirectional interaction between $X$ and $Y$ (the image and the latents), we can chain two reciprocal simplex attentions from $X$ to $Y$ and from $Y$ to $X$, obtaining the \textbf{\textit{duplex attention}}, which alternates computing $Y \coloneqq {u}^{a}(Y,X) $ and $X \coloneqq {u}^{d}(X,Y)$, such that each representation is refined in light of the other, integrating together bottom-up and top-down interactions. 

\subsubsection{Overall Architecture Structure}
\textbf{Vision-specific adaptations.} In the classic NLP transformer, each self-attention layer is followed by a feed-forward layer that processes each element independently, which can also be deemed a $1 \times 1$ convolution. Since our case pertains to images, we use instead a kernel size of $k=3$ after each attention operation. We further apply a Leaky ReLU nonlinearity after each convolution \citep{lrelu} and then upsample or downsmaple the features $X$, as part of the generator and discriminator respectively. To account for the features location within the image, we use a sinusoidal positional encoding \citep{transformer} along the horizontal and vertical dimensions for the visual features $X$, and trained positional embeddings for the set of latent variables $Y$.

\textbf{Model structure \& information flow.} Overall, the bipartite transformer is composed of a stack that alternates attention (simplex or duplex), convolution, and up- or down-sampling layers (see figure \ref{modelover}), starting from an initial $4 \times 4$ grid up to the desirable resolution for the generator, or progressing inversely for the distriminator. Conceptually, this structure fosters an interesting communication flow: rather than densely modeling interactions among all the pairs of pixels in the image, it supports adaptive long-range interaction between far away regions in a moderated manner, passing through a compact and global \textbf{\textit{latent bottleneck}}, that selectively gathers information from the entire input and distributes it back to the relevant regions. Intuitively, it can be viewed as analogous to the top-down and bottom-up notions discussed in section \ref{intro}, as information is propagated in the two directions, both from the local pixel to the global high-level representation and vice versa. 


\textbf{Computational efficiency.} We note that both the simplex and the duplex attention operations enjoy a bilinear efficiency of $\mathcal{O}(mn)$ thanks to the network's bipartite structure that considers all element pairs from $X$ and $Y$. Since, as we see below, we maintain $Y$ to be of a fairly small size, choosing $m$ in the range of 8--32, this compares favorably to the prohibitive $\mathcal{O}(n^2)$ complexity of self attention, which impedes its applicability to high-resolution images. 

\clearpage

\begin{figure*}[ht]
\centering
\subfloat{\includegraphics[width=0.2\linewidth]{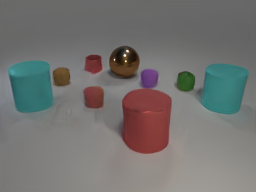}}
\subfloat{\includegraphics[width=0.2\linewidth]{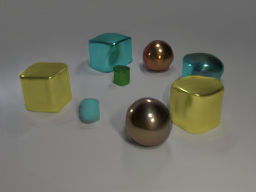}}
\subfloat{\includegraphics[width=0.2\linewidth]{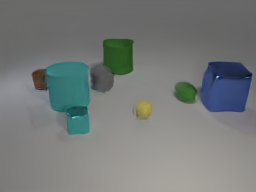}}
\subfloat{\includegraphics[width=0.2\linewidth]{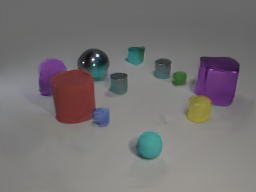}}
\subfloat{\includegraphics[width=0.2\linewidth]{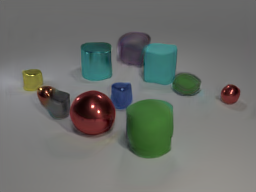}}
\vspace*{-10pt}
\subfloat{\includegraphics[width=0.2\linewidth]{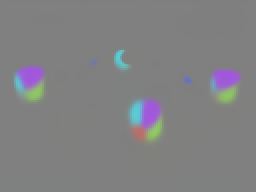}}
\subfloat{\includegraphics[width=0.2\linewidth]{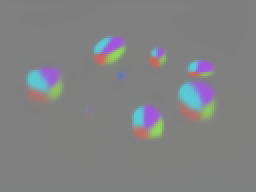}}
\subfloat{\includegraphics[width=0.2\linewidth]{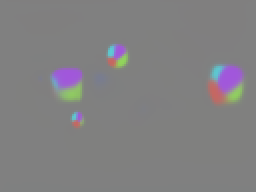}}
\subfloat{\includegraphics[width=0.2\linewidth]{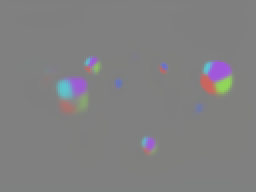}}
\subfloat{\includegraphics[width=0.2\linewidth]{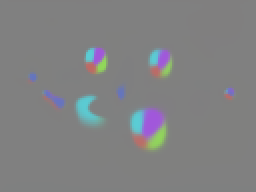}}
\vspace*{-10pt}
\subfloat{\includegraphics[width=0.2\linewidth]{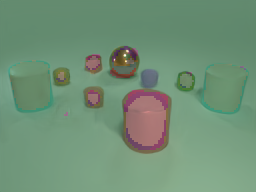}}
\subfloat{\includegraphics[width=0.2\linewidth]{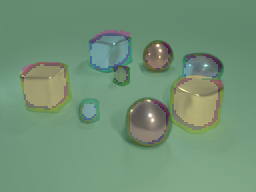}}
\subfloat{\includegraphics[width=0.2\linewidth]{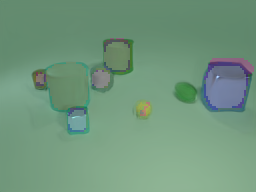}}
\subfloat{\includegraphics[width=0.2\linewidth]{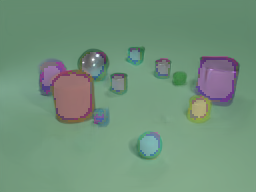}}
\subfloat{\includegraphics[width=0.2\linewidth]{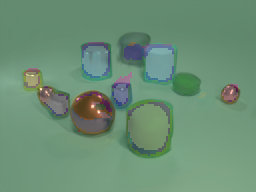}}
\vspace*{-5pt}
\subfloat{\includegraphics[width=0.2\linewidth]{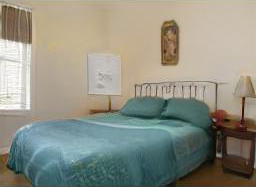}}
\subfloat{\includegraphics[width=0.2\linewidth]{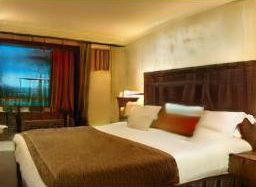}}
\subfloat{\includegraphics[width=0.2\linewidth]{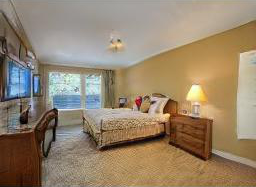}}
\subfloat{\includegraphics[width=0.2\linewidth]{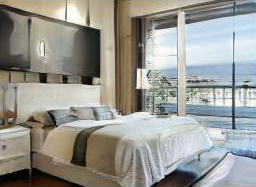}}
\subfloat{\includegraphics[width=0.2\linewidth]{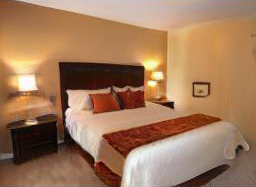}}
\vspace*{-10pt}
\subfloat{\includegraphics[width=0.2\linewidth]{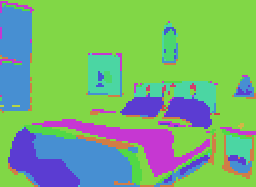}}
\subfloat{\includegraphics[width=0.2\linewidth]{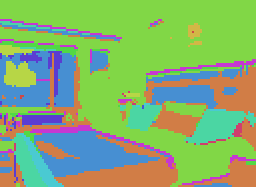}}
\subfloat{\includegraphics[width=0.2\linewidth]{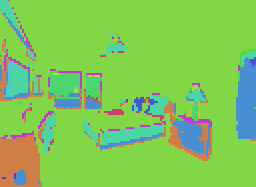}}
\subfloat{\includegraphics[width=0.2\linewidth]{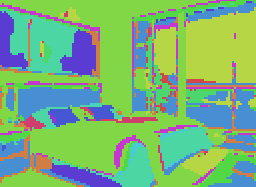}}
\subfloat{\includegraphics[width=0.2\linewidth]{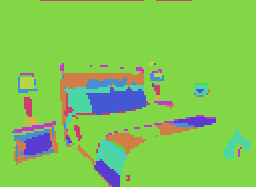}}
\vspace*{-5pt}
\subfloat{\includegraphics[width=0.2\linewidth]{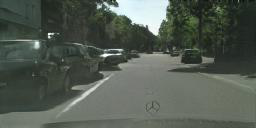}}
\subfloat{\includegraphics[width=0.2\linewidth]{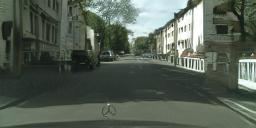}}
\subfloat{\includegraphics[width=0.2\linewidth]{images/final/nl/duplex/fakes00655208.png}}
\subfloat{\includegraphics[width=0.2\linewidth]{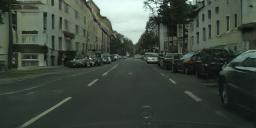}}
\subfloat{\includegraphics[width=0.2\linewidth]{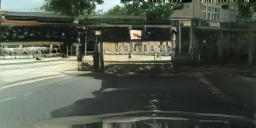}}
\vspace*{-10pt}
\subfloat{\includegraphics[width=0.2\linewidth]{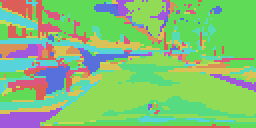}}
\subfloat{\includegraphics[width=0.2\linewidth]{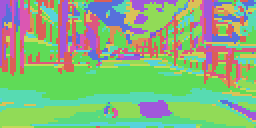}}
\subfloat{\includegraphics[width=0.2\linewidth]{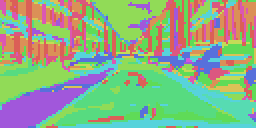}}
\subfloat{\includegraphics[width=0.2\linewidth]{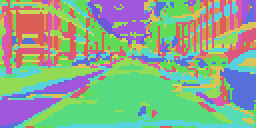}}
\subfloat{\includegraphics[width=0.2\linewidth]{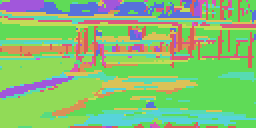}}
\vspace*{-4pt}
\caption{\textbf{Sample images and attention maps of lower and upper GANformer layers}, for the CLEVR, LSUN-Bedrooms, FFHQ and Cityscapes datasets. The colors in the attention maps correspond to the assignment between the image regions and the latent variables that control them. For the CLEVR dataset, we can see multiple attention maps produced by different layers of the model, revealing how the role of the latent variables changes at different stages of the generation -- while they correspond to an  \textit{\textbf{instance segmentation}} as the layout of the scene is being formed in the \textbf{early low-resolution layers}, they behave similarly to a \textit{\textbf{surface normal}} in the \textbf{upper high-resolution layers} of the generator. We see similar progression from a coarser to finer pattern of attention for the FFHQ dataset.}
\label{layer_attmaps}
\end{figure*} 

\clearpage

\begin{table*}[!ht] 
\caption{\textbf{Comparison between the GANformer and competing methods for image synthesis.} We evaluate the models along commonly used metrics of FID, Precision and Recall scores. FID is well-received as a reliable indication of image fidelity and diversity, while Precision and Recall measure the similarity between the generated and natural distributions. Metrics are computed over 50k samples.} 
\vspace*{5.5pt}
\label{table1}
\centering
\scriptsize
\begin{tabular}{lcccccccc}
\rowcolor{Blue1}
\scriptsize & {\color{Blue} \textbf{CLEVR}} &  &  &  & {\color{Blue} \textbf{LSUN-Bedrooms}} &  &  &  \\ 
\rowcolor{Blue1}
\textbf{Model} & FID $\downarrow$ & IS $\uparrow$ & Precision $\uparrow$ & Recall $\uparrow$ & FID $\downarrow$ & IS $\uparrow$ & Precision $\uparrow$ & Recall $\uparrow$ \\
\scriptsize GAN & 25.02 & 2.17 & 21.77 & 16.76 & 12.16 & 2.66 & 52.17 & 13.63 \\
\scriptsize k-GAN & 28.29 & 2.21 & 22.93 & 18.43 & 69.90 & 2.41 & 28.71 & 3.45 \\
\scriptsize SAGAN & 26.04 & 2.17 & 30.09 & 15.16 & 14.06 & 2.70 & 54.82 & 7.26 \\
\scriptsize StyleGAN2 & 16.05 & 2.15 & 28.41 & 23.22 & 11.53 & \textbf{2.79} & 51.69 & 19.42 \\
\rowcolor{Blue2}
\scriptsize \textbf{GANformer$_\text{s}$} & 10.26 & \textbf{2.46} & 38.47 & 37.76 & 8.56 & 2.69 & 55.52 & 22.89 \\
\rowcolor{Blue2}
\scriptsize \textbf{GANformer$_\text{d}$} & \textbf{9.17} & 2.36 & \textbf{47.55} & \textbf{66.63} & \textbf{6.51} & 2.67 & \textbf{57.41} & \textbf{29.71} \\
\rowcolor{Gray}
\tiny  &  &  &  &  &  &  &  &  \\
\rowcolor{Blue1}
\scriptsize & {\color{Blue} \textbf{FFHQ}} &  &  &  & {\color{Blue} \textbf{Cityscapes}} &  &  &  \\ 
\rowcolor{Blue1}
\textbf{Model} & FID $\downarrow$ & IS $\uparrow$ & Precision $\uparrow$ & Recall $\uparrow$ & FID $\downarrow$ & IS $\uparrow$ & Precision $\uparrow$ & Recall $\uparrow$ \\
\scriptsize GAN & 13.18 & 4.30 & 67.15 & 17.64 & 11.57 & 1.63 & 61.09 & 15.30  \\
\scriptsize k-GAN & 61.14 & 4.00 & 50.51 & 0.49 & 51.08 & 1.66 & 18.80 & 1.73 \\
\scriptsize SAGAN & 16.21 & 4.26 & 64.84 & 12.26 & 12.81 & 1.68 & 43.48 & 7.97 \\
\scriptsize StyleGAN2 & 9.24 & 4.33 & 68.61 & \textbf{25.45} & 8.35 & \textbf{1.70} & 59.35 & 27.82 \\
\rowcolor{Blue2}
\scriptsize \textbf{GANformer$_\text{s}$} & 8.12 & \textbf{4.46} & \textbf{68.94} & 10.14 & 14.23 & 1.67 & \textbf{64.12} & 2.03 \\
\rowcolor{Blue2}
\scriptsize \textbf{GANformer$_\text{d}$} & \textbf{7.42} & 4.41 & 68.77 & 5.76 & \textbf{5.76} & 1.69 & 48.06 & \textbf{33.65} \\
\rowcolor{Blue2}
\end{tabular}
\vspace*{-5pt}
\end{table*}

\subsection{The Generator and Discriminator Networks}
\label{gan}
\vspace*{-2.5pt}

Our generator and discriminator networks follow the general design of prior work \cite{stylegan, stylegan2}, with the key difference of incorporating the novel bipartite attention layers instead of the single-latent modulation that characterizes earlier models: Commonly, a generator network consists of a multi-layer CNN that receives a randomly sampled vector $z$ and transforms it into an image. The popular StyleGAN approach departs from this design and, instead, introduces a feed-forward \textit{mapping network} that outputs an intermediate vector $w$, which in turn interacts directly with each convolution through the \textit{synthesis network}, globally modulating the feature maps' statistics at every layer. 

Effectively, this approach attains \textit{layer-wise decomposition} of visual properties, allowing StyleGAN to control global aspects of the picture such as the pose, lighting conditions or color scheme, in a coherent manner over the entire image. But while StyleGAN successfully disentangles global attributes, it is more limited in its ability to perform \textit{spatial decomposition}, as it provides no direct means to control the style of localized regions within the generated image. 


The bipartite transformer offers a solution to accomplish this objective. Instead of modulating the style of all features globally, we use instead our new attention layer to perform adaptive region-wise modulation. As shown in figure \ref{modelover} (right), we split the latent vector $z$ into $k$ components, $z = [z_1,....,z_k]$ and, as in StyleGAN, pass each of them through a shared mapping network, obtaining a corresponding set of intermediate latent variables $Y=[y_1,...,y_k]$. Then, during synthesis, after each CNN layer of the generator, we let the feature map $X$ and latents $Y$ play the roles of the two element groups, mediating their interaction through our new attention layer -- either simplex or duplex. 

This setting thus allows for a \textbf{flexible and dynamic style modulation at the level of the region}. Since soft attention tends to group elements based on their \textbf{proximity and content similarity}, we see how the transformer architecture naturally fits into the generative task and proves useful in the visual domain, allowing the model to exercise finer control in modulating local semantic regions. As we see in section \ref{exps}, this capability turns out to be especially useful in modeling highly-structured scenes. 



As to the loss function, optimization and training configurations, we adopt the settings and techniques used by StyleGAN2 \citep{stylegan2}, including in particular style mixing, stochastic variation, exponential moving average for weights, and a non-saturating logistic loss with lazy R1 regularization\footnote{In the prior version of the paper and in earlier stages of the model development, we explored incorporating the bipartite attention to both the generator and the discriminator, in order to allow both components make use of long-range interactions. However, in ablation experiments we observed that applying attention to the generator only allows for stronger results, and so we have updated the paper accordingly.}. 

\subsection{Summary} 
\label{discussion}
To recapitulate the discussion above, the GANformer successfully unifies the GAN and Transformer architectures for the task of scene generation. Compared to traditional GANs and transformers, it introduces multiple key innovations:
\begin{itemize}
  \item \textbf{Compositional Latent Space} with multiple variables that coordinate through attention to produce the image cooperatively, in a manner that matches the inherent compositionality of natural scenes.
  \item \textbf{Bipartite Structure} that balances between expressiveness and efficiency, modeling long-range dependencies while maintaining linear computational costs.
  \item \textbf{Bidirectional Interaction} between the latents and the visual features, which allows the refinement and interpretation of each in light of the other.
  \item \textbf{Multiplicative Integration} rule to impact the features' visual style more flexibly, akin to StyleGAN but in contrast to the classic transformer network.
\end{itemize}
As we see in the following section, the combination of these design choices yields a strong architecture that demonstrates high efficiency, improved latent space disentanglement, and enhanced transparency of the generative process.

\section{Experiments}
\label{exps}

We investigate the GANformer through a suite of experiments that study its quantitative performance and qualitative behavior. As we will see below, the GANformer achieves state-of-the-art results, successfully producing high-quality images for a varied assortment of datasets: FFHQ for human faces \citep{stylegan}, the CLEVR dataset for multi-object scenes \citep{clevr}, and the LSUN-Bedrooms \citep{lsun} and Cityscapes \citep{cityscapes} datasets for challenging indoor and outdoor scenes. Notably, it even attains state-of-the-art FID scores for the challenging and highly-structured COCO dataset. 

Further analysis we conduct in sections \ref{data}, \ref{comp} and \ref{dis} provides evidence for multiple favorable properties the GANformer posses, including better data-efficiency, enhanced transparency, and stronger disentanglement than prior approaches. Section \ref{dvrs} then quantitatively assesses the network's semantic coverage of the natural image distribution for the CLEVR dataset, while ablation and variation studies at section \ref{ablt} empirically validate the necessity of each of the model's design choices. Taken altogether, our evaluation offers solid evidence for the GANformer's effectiveness and efficacy in modeling compsitional images and scenes. 

We compare our network with several  approaches, including both baselines and leading models for image synthesis: (1)~A baseline GAN \citep{gan} that follows the typical convolutional architecture\footnote{In the baseline GAN, we input the noise through the network's stem instead of through weight modulation.}; (2)~StyleGAN2 \citep{stylegan2}, where a single global latent interacts with the evolving image by modulating its global style; (3)~SAGAN \citep{sagan}, which performs self attention across all feature pairs in low-resolution layers of the generator and the discriminator; and (4)~k-GAN \citep{kgan} that produces $k$ separated images, which are then blended through alpha-composition. 

%
To evaluate all models under comparable training conditions, model size, and optimization scheme, we implement them all within our public codebase, which extends the official StyleGAN repository. All models have been trained with images of $256 \times 256$ resolution and for the same number of training steps, roughly spanning a week on 2 NVIDIA V100 GPUs per model (or equivalently 3-4 days using 4 GPUs). For the GANformer, we select $k$ -- the number of latent variables, from the range of 8--32. Note that increasing the value of $k$ does not translate to an increased overall latent dimension, and we rather keep it equal across models. See section \ref{impl} for further implementation details, hyperparameter settings and training configurations. 




\begin{figure*}[!t]
\centering
\subfloat{\includegraphics[width=0.25\linewidth]{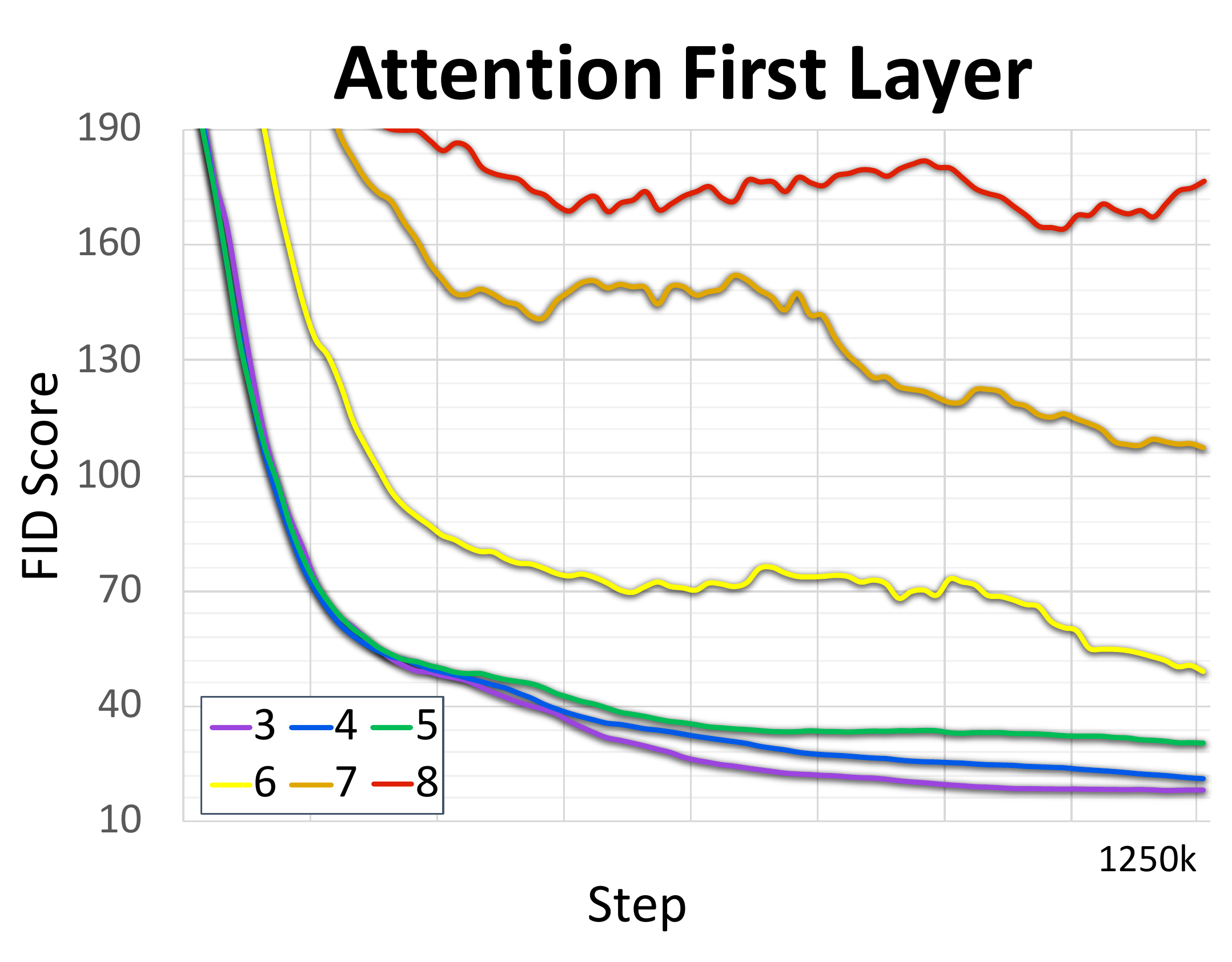}}
\hfill
\subfloat{\includegraphics[width=0.25\linewidth]{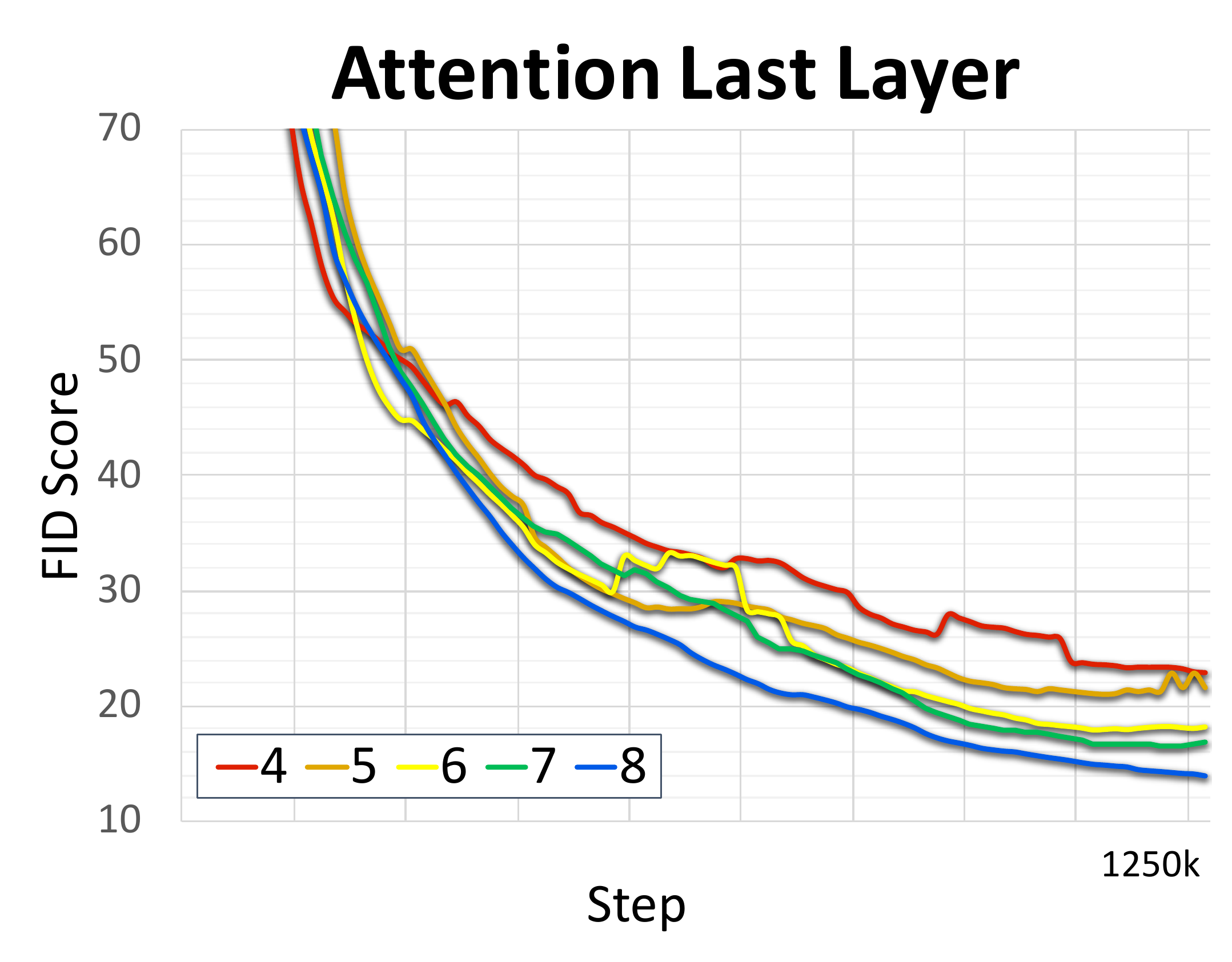}}
\hfill
\subfloat{\includegraphics[width=0.25\linewidth]{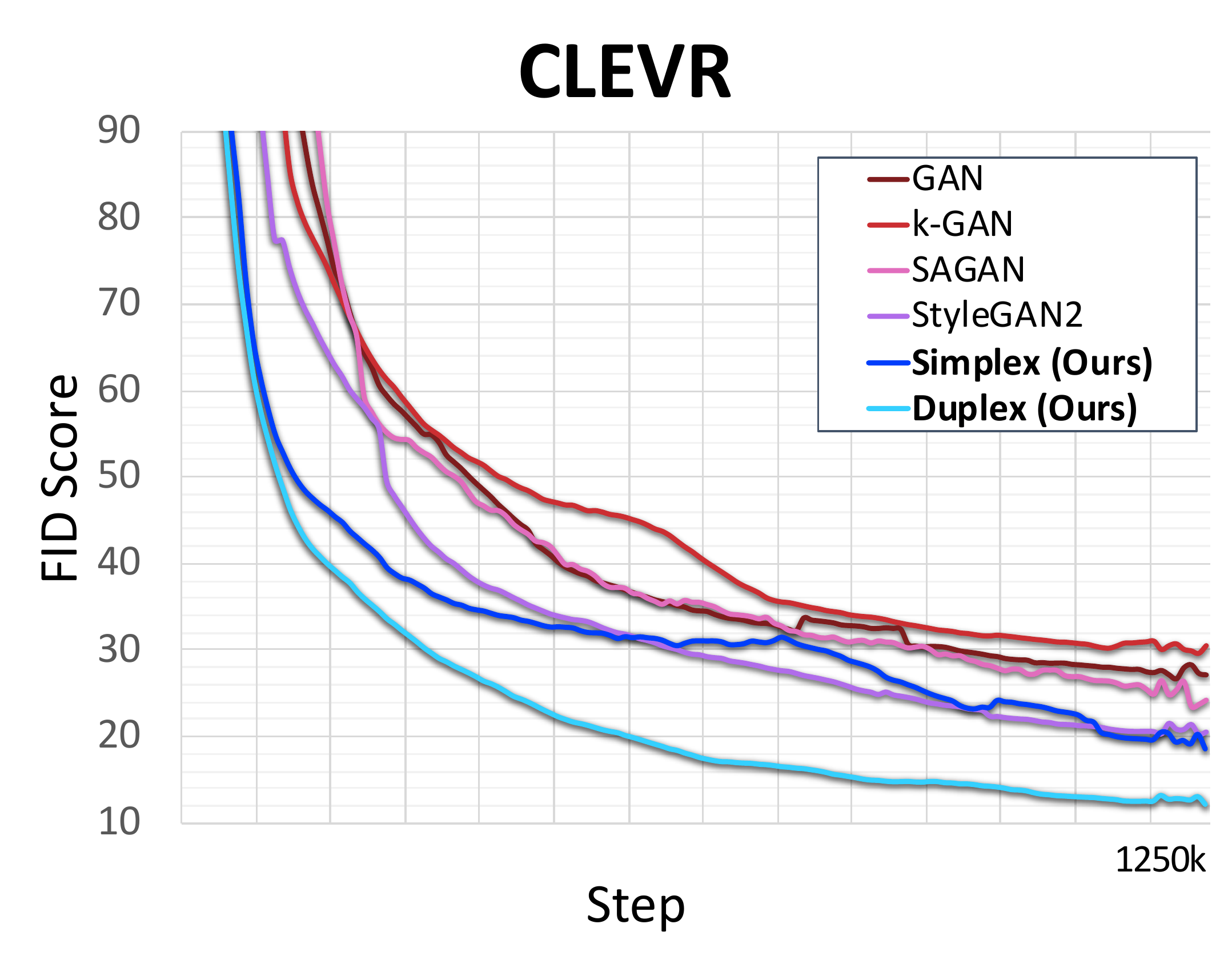}}
\hfill
\subfloat{\includegraphics[width=0.25\linewidth]{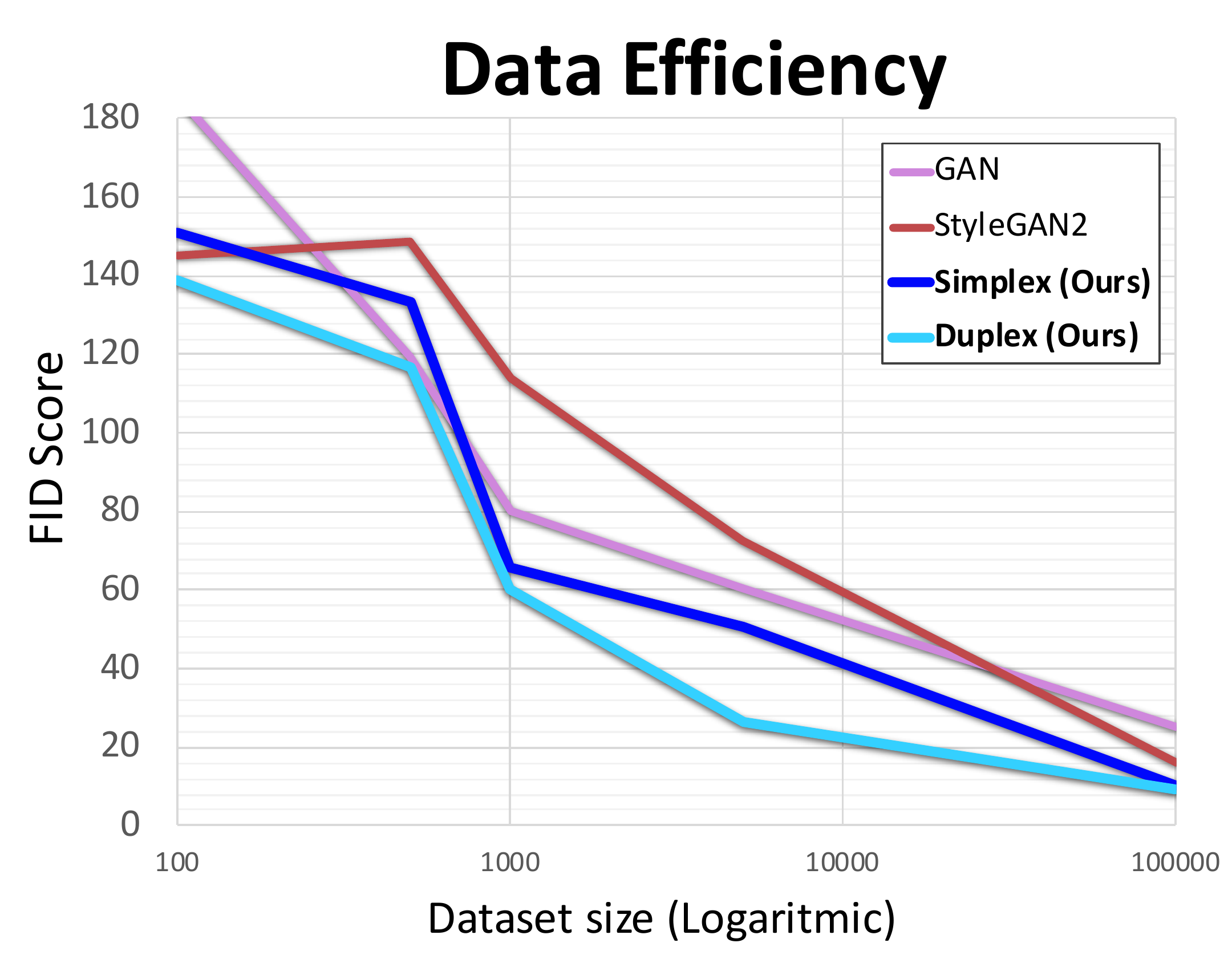}}
\vspace*{-0.8mm}
\caption{\textbf{From left to right: (1-2) Learning Performance} as a function of the earliest and latest layers that the bipartite attention is applied to. The more layers attention is used through, the better the model's performance gets and the faster it learns, confirming the effectiveness of our approach. \textbf{(3) Learning Curves} for the GANformer vs. competing approaches, demonstrating its fast learning. \textbf{(4): Data-Efficiency} for CLEVR: performance as a function of the training set size.} 
\label{plots}
\vspace*{-12pt}
\end{figure*}

As shown in table \ref{table1}, our model matches or outperforms prior work, achieving substantial gains in terms of FID score, which correlates with image quality and diversity \citep{fid}, as well as other commonly used metrics such as Precision and Recall (P{\&}R)\footnote{Note that while the StyleGAN paper \citep{stylegan2} reports lower FID scores for FFHQ and LSUN-Bedrooms, they are obtained by training for 5-7 times longer than our experiments (specifically, they train for up to 17.5 million steps, producing 70M samples and demanding over 90 GPU-days). To comply with a reasonable compute budget, we equally reduced the training duration for all models in our evaluation, maintaining the same number of steps.}. As could be expected, we obtain the least gains for the FFHQ human faces dataset, where naturally there is relatively lower diversity in image layout. On the flip side, most notable are the significant improvements in performance for CLEVR, where our approach successfully lowers FID scores from 16.05 to 9.17, as well as LSUN-Bedrooms, where the GANformer nearly halves the FID score from 11.53 to 6.51, being trained for equal number of steps. These findings suggest that the GANformer is particularly adept at modeling scenes of high compositionality (CLEVR) or layout diversity (LSUN-Bedrooms). Comparing between the Simplex and Duplex Attentions further reveals the strong benefits of integrating the reciprocal bottom-up and top-down processes together. 

\subsection{Data and Learning Efficiency}
\label{data} 
We examine the learning curves of our and competing models (figure \ref{plots}, (3)) and inspect samples of generated images at different stages of the training (figure \ref{training}). These results both indicate that our model learns significantly faster than competing approaches. In the case of CLEVR, it produces high-quality images in approximately 3-times less training steps than the second-best approach. To further explore the GANformer's learning aptitude, we perform experiments where we reduce the size of the dataset each model (and specifically, its discriminator) is exposed to during training to varying degrees (figure \ref{plots}, (4)). These results similarly validate the model's superior data-efficiency, especially where as few as 1k images are provided for training. 



\vspace*{-2.65pt}
\subsection{Transparency \& Compositionality} 
\label{comp}



To gain more insight into the model's internal representation and its underlying generative process, we visualize the attention distributions produced by the GANformer as it synthesizes new images. Recall that at each layer of the generator, it casts attention between the $k$ latent variables and the evolving spatial features of the generated image. 

As illustrated by figures \ref{imgs} and \ref{layer_attmaps}, the latent variables tend to attend to coherent visual regions in terms of proximity and content similarity. Figure \ref{layer_attmaps} provides additional attention maps computed by the model in various layers, showing how it behaves distinctively in different stages of the generation process. The visualizations imply that the latents carry a semantic sense, capturing objects, visual entities or other constituent components of the synthesized scenes. These findings can thereby attest to an enhanced compositionality that our model acquires through its multi-latent structure. Whereas prior work uses a single monolithic latent vector to account for the whole scene and modulate features at a global scale only, our design lets the GANformer exercise finer control that impacts features at the object granularity, while leveraging the use of attention to make its internal representations more structured and transparent.  

To quantify the compositionality exhibited by the model, we use a pre-trained detector \citep{detectron2} to produce  segmentations for a set of generated scenes, in order to measure the correlation between the attention cast by the latents with various semantic classes. Figure \ref{corr}shows the classes that have the highest correlation with respect to the latent variables, indicating that different latents indeed coherently attend to semantic concepts such as windows, pillows, sidewalks or cars, as well as background regions like carpets, ceiling, and walls. This illustrates how the multiple latents are effectively used to semantically decompose the scene generation task. 
  


\begin{table}[t]
\vspace*{-6pt}
\caption{\textbf{Chi-Square Statistics} for CLEVR generated scenes, based on 1k samples. Images were processed by a pre-trained object detector, identifying objects and semantic attributes, to compute the properties' distribution across the generated scenes.} 
\vspace*{8pt}

\label{dist}
\centering
\scriptsize
\begin{tabular}{lcccc}
\rowcolor{Blue1}
& \textbf{GAN} & \textbf{StyleGAN} & \textbf{GANformer$_s$} & \textbf{GANformer$_d$} \\
\scriptsize \textbf{Object Area} & 0.038 & 0.035 & 0.045 & \textbf{0.068} \\
\rowcolor{Blue2}
\scriptsize \textbf{Object Number} & 2.378 & 1.622 & 2.142 & \textbf{2.825} \\
\scriptsize \textbf{Co-occurrence} & \textbf{13.532} & 9.177 & 9.506 & 13.020 \\
\rowcolor{Blue2}
\scriptsize \textbf{Shape} & 1.334 & 0.643 & 1.856 & \textbf{2.815} \\
\scriptsize \textbf{Size} & 0.256 & 0.066 & 0.393 & \textbf{0.427} \\
\rowcolor{Blue2}
\scriptsize \textbf{Material} & 0.108 & 0.322 & 1.573 & \textbf{2.887} \\
\scriptsize \textbf{Color} & 1.011 & 1.402 & 1.519 & \textbf{3.189} \\
\rowcolor{Blue2}
\scriptsize \textbf{Class} & 6.435 & 4.571 & 5.315 & \textbf{16.742} \\
\end{tabular}
\vspace*{-14pt}
\end{table}

\begin{table}[t]
\vspace*{-2pt}
\caption{\textbf{Disentanglement metrics (DCI and modularity)}, which asses the Disentanglement, Completeness  Informativeness, and Modularity of the latent representations, effectively measuring their correspondance to visual attributes in the out images, computed over 1k CLEVR samples. The GANformer achieves the strongest results compared to competing approaches.}
\vspace*{8pt}
\label{disen}
\centering
\scriptsize
\begin{tabular}{lcccc}
\rowcolor{Blue1}
\scriptsize & \textbf{GAN} & \textbf{StyleGAN} & \textbf{GANformer$_s$} & \textbf{GANformer$_d$} \\
\rowcolor{Blue2}
\scriptsize \textbf{Disentanglement} & 0.126 & 0.208 & 0.556 & \textbf{0.768} \\
\scriptsize \textbf{Modularity} & 0.631 & 0.703 & 0.891 & \textbf{0.952} \\
\rowcolor{Blue2}
\scriptsize \textbf{Completeness} & 0.071 & 0.124 & 0.195 & \textbf{0.270} \\
\scriptsize \textbf{Informativeness} & 0.583 & 0.685 & 0.899 & \textbf{0.972} \\
\rowcolor{Blue2}
\scriptsize \textbf{Informativeness'} & 0.434 & 0.332 & 0.848 & \textbf{0.963} \\
\end{tabular}
\vspace*{-13pt}
\end{table}

\subsection{Disentanglement}
\label{dis}

We consider the DCI and Modularity metrics commonly used in the disentanglement literature \citep{dci,modularity} to provide more evidence for the beneficial impact our architecture has on the model's internal representation. These metrics asses the Disentanglement, Completeness, Informativeness and Modularity of a given representation, essentially evaluating the degree to which there is a 1-to-1 correspondence between latent factors and global image attributes. To obtain the attributes, we consider the area size of each semantic class (e.g. cubes, spheres, floor), predicted by a pre-trained segmentor, and use them as the output response features for measuring the latent space disentanglement, computed over 1k images. We follow the protocol proposed by Wu et al. \citep{stylespace} and present the results in table \ref{disen}. This analysis confirms that the GANformer's latent representations enjoy higher disentanglement compared to competing approaches. 

\subsection{Image Diversity}
\label{dvrs}

A major advantage of compositional representations is that they can support combinatorial generalization -- a key foundation of human intelligence \cite{relational}. Inspired by this observation, we measure this property in the context of visual synthesis of multi-object scenes. We use a pre-trained object detector on generated CLEVR scenes, to extract the objects and properties within each sample. We then compute Chi-Square statistics on the sample set to determine the degree to which each model manages to cover the natural uniform distribution of CLEVR images. Table \ref{dist} summarizes the results, where we can see that our model obtains better scores across almost all the semantic properties of the scenes distribution. These metrics complement the common FID and PR scores as they emphasize structure over texture, or semantics over perceptual appearance, focusing on object existence, arrangement and local properties, and thereby substantiating further the model's compositionality. %

\subsection{Ablation Studies}
\label{ablt} 
To validate the usefulness of bipartite attention, we conduct ablation studies, where we vary the index of the earliest and latest layers of the generator network to which attention is incorporated. As indicated by figure \ref{plots} (1-2), the earlier (or lower resolution) attention begins being applied, the better the model's performance and the faster it learns. The same goes for the latest layer to apply attention to -- as attention can especially contribute in high-resolutions, which benefit the most from long-range interactions. These studies provide a validation for the effectiveness of our approach in enhancing generative scene modeling. 

\section{Conclusion}
\label{conclusion}
We have introduced the GANformer, a novel and efficient bipartite transformer that combines top-down and bottom-up interactions, and explored it for the task of generative modeling, achieving strong quantitative and qualitative results that attest to the model robustness and efficacy. The GANformer fits within the general philosophy that aims to incorporate stronger inductive biases into neural networks to encourage desirable properties such as transparency, data-efficiency and compositionality -- properties which are at the core of human intelligence, serving as the basis for our capacity to plan, reason,  learn, and imagine. While our work focuses on visual synthesis, we note that the bipartite transformer is a general-purpose model, and expect it may be found useful for other tasks in both vision and language. Overall, we hope that our work will help progressing further in our collective search to bridge the gap between the intelligence of humans and machines. 

\section{Acknowledgments}
\label{ack}
We are grateful to Stanford HAI for the generous computational resources provided through Amazon AWS cloud credits. I also wish to thank Christopher D. Manning for the fruitful discussions and constructive feedback in developing the bipartite transformer, especially when we explored it for language representation, as well as for the kind financial support he provided that allowed this work to happen.  

\bibliography{example_paper}
\bibliographystyle{neurips_2019}

\cleardoublepage
\newpage

\appendix

\section*{Supplementary Material}

In the following, we provide additional experiments and visualizations for the GANformer model. First, we present in figures \ref{training} and \ref{sota} a comparison of sample images produced by the GANformer and a set of baseline models, over the course of the training and after convergence respectively. Section \ref{impl} specifies the implementation details, optimization scheme and training configuration of the model. In section \ref{spc} and figure \ref{corr}, we evaluate the spatial compositionality of the GANformer's attention mechanism, shedding light upon the roles of the different latent variables. 


\section{Implementation and Training Details}
\label{impl}
To evaluate all models under comparable conditions of training configuration, model size, and optimization details, we implement them all within the TensorFlow codebase introduced by the StyleGAN authors \citep{stylegan}. See table \ref{hyperparams} for particular settings of the GANformer and table \ref{paramsnum} for comparison of model sizes. 

In terms of the loss function, optimization and training configuration, we adopt the settings and techniques used in the StyleGAN2 model \citep{stylegan2}, including in particular style mixing, Xavier Initialization, stochastic variation, exponential moving average for weights, and a non-saturating logistic loss with lazy a R1 regularization. We use Adam optimizer with batch size of 32 (4 $\times$ 8 using gradient accumulation), equalized learning rate of $0.001$, $\beta_1 = 0.0$ and $\beta_2 = 0.99$ as well as leaky ReLU activations with $\alpha = 0.2$, bilinear filtering in all up/downsampling layers and minibatch standard deviation layer at the end of the discriminator. The mapping layer of the generator consists of 8 layers, and ResNet connections are used throughout the model, for the mapping network, synthesis network and discriminator. 

We train all models on images of $256 \times 256$ resolution, padded as necessary. The CLEVR dataset consists of 100k images, the FFHQ has 70k images, Cityscapes has overall about 25k images and LSUN-Bedrooms has 3M images. The images in the Cityscapes and FFHQ datasets are mirror-augmented to increase the effective training set size. All models have been trained for the same number of training steps, roughly spanning a week on 2 NVIDIA V100 GPUs per model. 

\section{Spatial Compositionality}
\label{spc}
\begin{figure}[ht]
\centering
\subfloat{\includegraphics[width=0.4\linewidth]{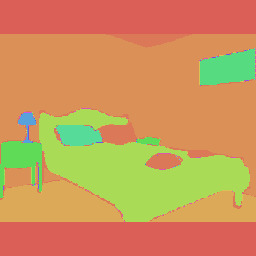}}
\subfloat{\includegraphics[width=0.48\linewidth]{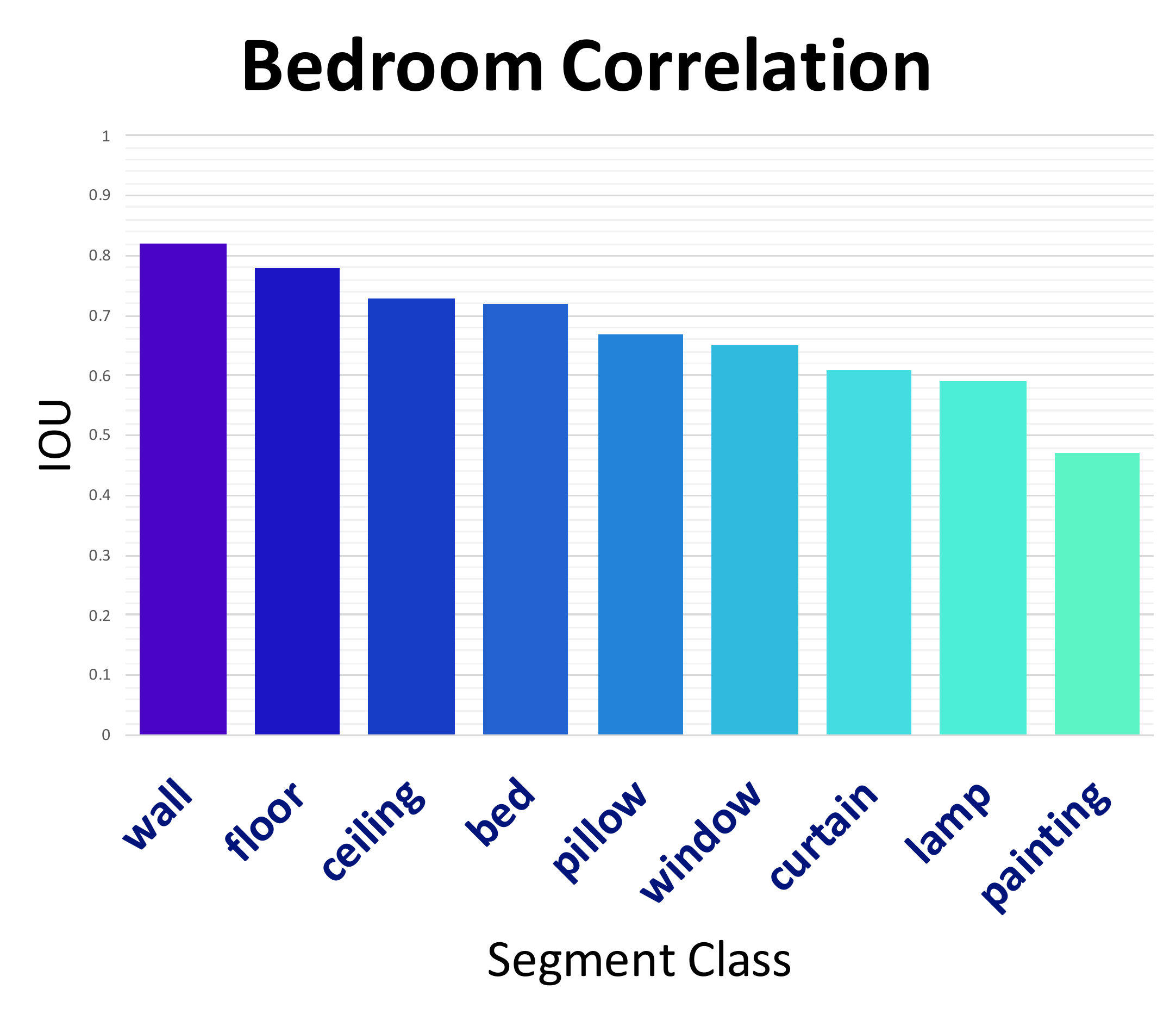}}
\vspace*{-10pt}
\subfloat{\includegraphics[width=0.4\linewidth]{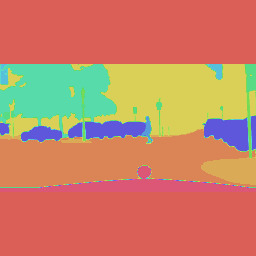}}
\subfloat{\includegraphics[width=0.48\linewidth]{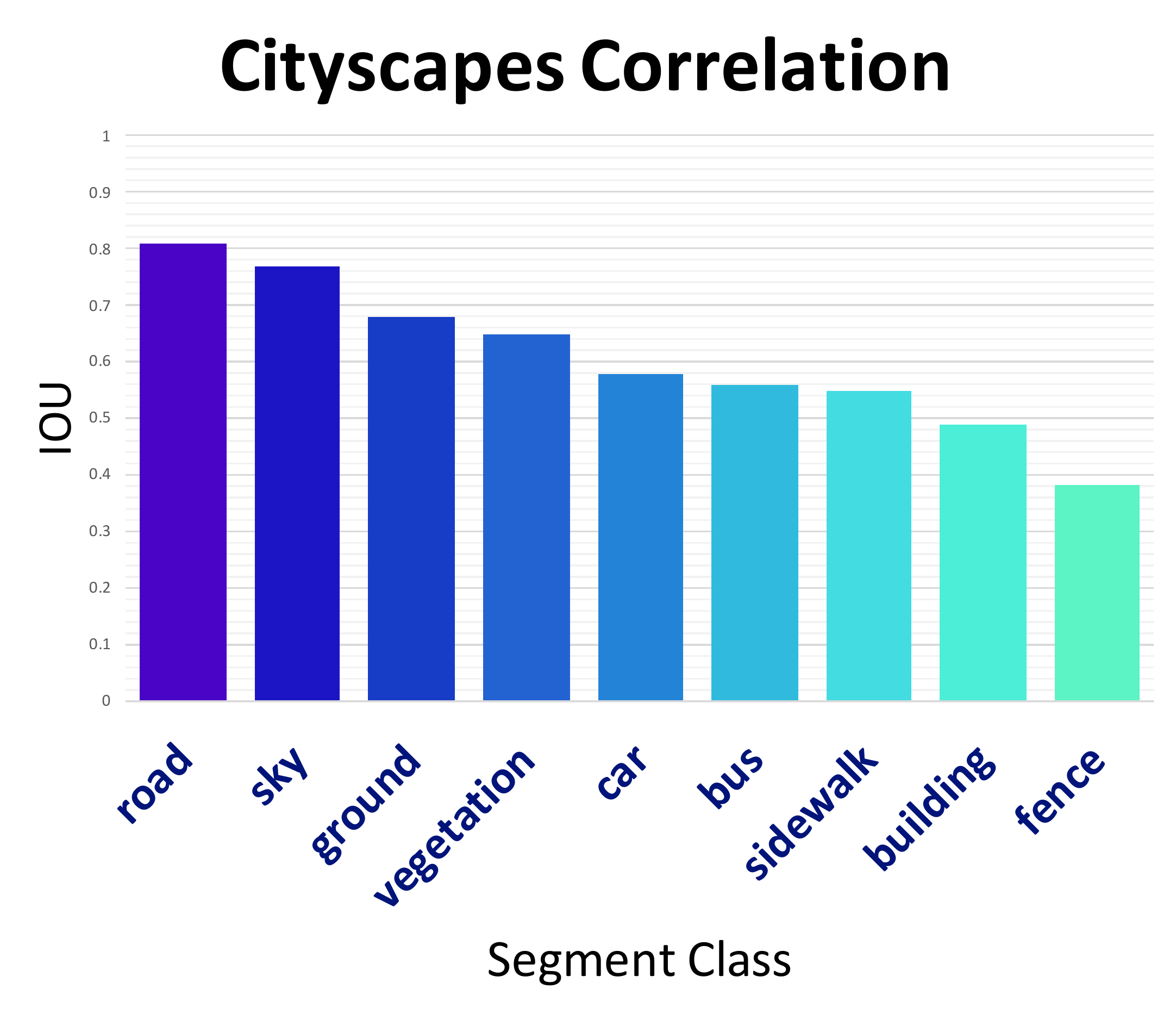}}
\vspace*{-2mm}
\caption{\textbf{Spatial compositionality.} Correlation between attention maps and semantic segments, computed over 1k samples. Results are presented for the LSUN-Bedrooms and Cityscapes.}
\vspace*{-8pt}
\label{corr}
\end{figure}

To quantify the compositionality level exhibited by the model, we employ a pre-trained segmentor to produce semantic segmentations for the synthesized scenes, and use them to measure the correlation between the attention cast by the latent variables and the various semantic classes. We derive the correlation by computing the maximum intersection-over-union between a class segment and the attention segments produced by the model in the different layers. The mean of these scores is then taken over a set of 1k images. Results presented in figure \ref{corr} for the LSUN-Bedrooms and Cityscapes datasets, showing semantic classes which have high correlation with the model attention, indicating it decomposes the image into semantically-meaningful segments of objects and entities.

\vspace*{-5pt}
\begin{table}[h]
\caption{\textbf{Hyperparameter choices}. The latents number (each variable is multidimensional) is chosen based on performance among $\{8, 16, 32, 64\}$. The overall latent dimension is chosen among $\{128, 256, 512\}$ and is then used both for the GANformer and the baseline models. The R1 regularization factor $\gamma$ is chosen among $\{1, 10, 20, 40, 80, 100\}$.}
\vspace*{8pt}
\label{hyperparams}
\centering
\scriptsize
\begin{tabular}{lcccc}
\rowcolor{Blue1}
& \textbf{FFHQ} & \textbf{CLEVR} & \textbf{Cityscapes} & \textbf{Bedroom} \\
\scriptsize \# \textbf{Latent var} & 8 & 16 & 16 & 16 \\
\rowcolor{Blue2}
\scriptsize \textbf{Latent var dim} & 16 & 32 & 32 & 32 \\
\scriptsize \textbf{Latent overall dim} & 128 & 512 & 512 & 512 \\
\rowcolor{Blue2}
\scriptsize \textbf{R1 reg weight} ($\gamma$) & 10 & 40 & 20 & 100 \\
\end{tabular}
\end{table}

\begin{table}[h] 
\caption{\textbf{Model size} for the GANformer and competing approaches, computed given 16 latent variables and an overall latent dimension of 512. All models are comparable in size.}
\vspace*{8pt}
\label{paramsnum}
\centering
\scriptsize
\begin{tabular}{lcc}
\rowcolor{Blue1}
 & \# \textbf{G Params} & \# \textbf{D Params} \\
\scriptsize \textbf{GAN} & 34M & 29M \\
\rowcolor{Blue2}
\scriptsize \textbf{StyleGAN2} & 35M  & 29M \\
\scriptsize \textbf{k-GAN} & 34M & 29M \\
\rowcolor{Blue2}
\scriptsize \textbf{SAGAN} & 38M & 29M \\
\scriptsize \textbf{GANformer$_s$} & 36M & 29M \\
\rowcolor{Blue2}
\scriptsize \textbf{GANformer$_d$} & 36M & 29M \\
\end{tabular}
\end{table}

\begin{figure*}[t]
\centering
\setlength{\tabcolsep}{0pt} 
\renewcommand{\arraystretch}{0} 
\begin{tabular}{c c c c c c}
\rowcolor{white}

\textbf{{GAN  }} & \includegraphics[width=0.165\linewidth]{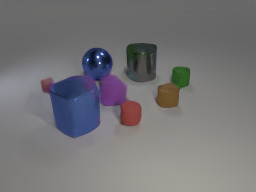} & \includegraphics[width=0.165\linewidth]{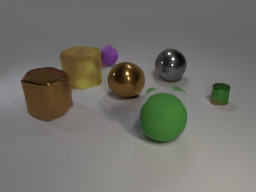} & \includegraphics[width=0.165\linewidth]{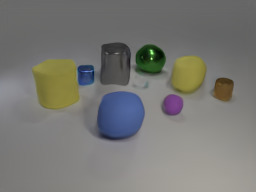} & \includegraphics[width=0.165\linewidth]{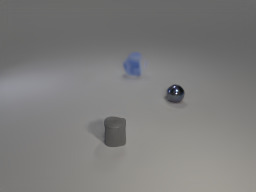} & \includegraphics[width=0.165\linewidth]{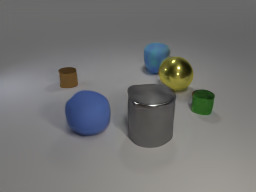} \\

 & \includegraphics[width=0.165\linewidth]{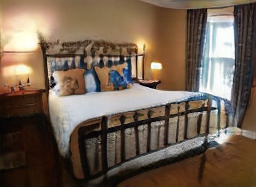} & \includegraphics[width=0.165\linewidth]{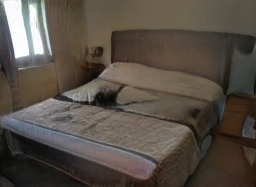} & \includegraphics[width=0.165\linewidth]{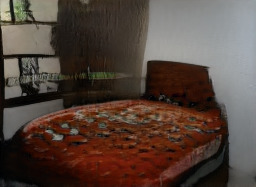} & \includegraphics[width=0.165\linewidth]{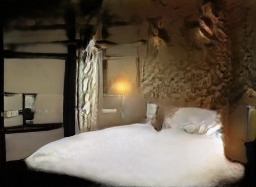} & \includegraphics[width=0.165\linewidth]{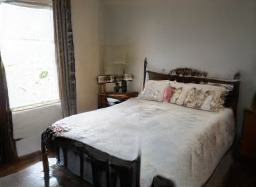} \\

\vspace*{12pt}

 & \includegraphics[width=0.165\linewidth]{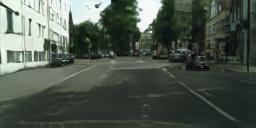} & \includegraphics[width=0.165\linewidth]{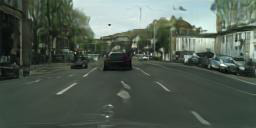} & \includegraphics[width=0.165\linewidth]{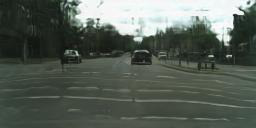} & \includegraphics[width=0.165\linewidth]{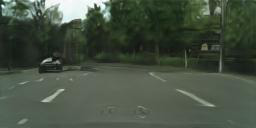} & \includegraphics[width=0.165\linewidth]{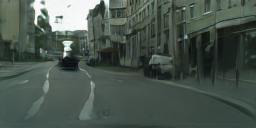} \\

\textbf{{StyleGAN2  }} & \includegraphics[width=0.165\linewidth]{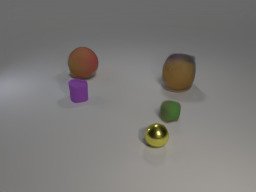} & \includegraphics[width=0.165\linewidth]{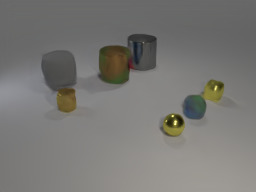} & \includegraphics[width=0.165\linewidth]{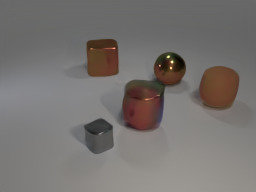} & \includegraphics[width=0.165\linewidth]{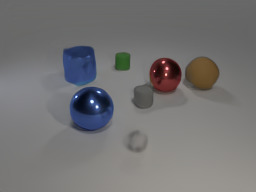} & \includegraphics[width=0.165\linewidth]{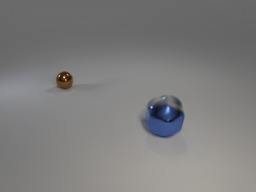} \\

 & \includegraphics[width=0.165\linewidth]{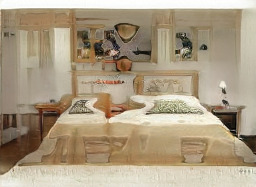} & \includegraphics[width=0.165\linewidth]{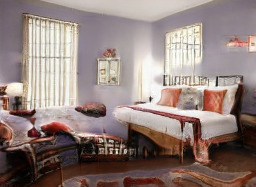} & \includegraphics[width=0.165\linewidth]{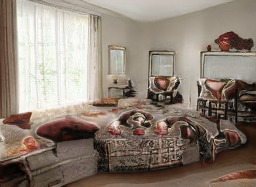} & \includegraphics[width=0.165\linewidth]{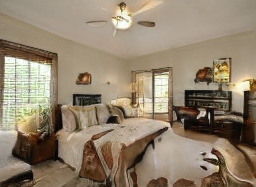} & \includegraphics[width=0.165\linewidth]{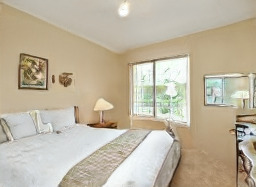} \\

\vspace*{12pt}

 & \includegraphics[width=0.165\linewidth]{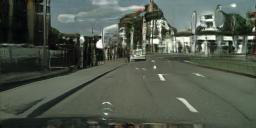} & \includegraphics[width=0.165\linewidth]{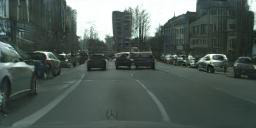} & \includegraphics[width=0.165\linewidth]{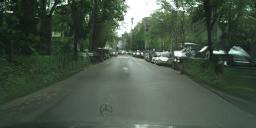} & \includegraphics[width=0.165\linewidth]{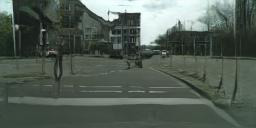} & \includegraphics[width=0.165\linewidth]{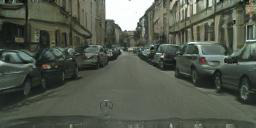} \\

\textbf{{k-GAN  }} & \includegraphics[width=0.165\linewidth]{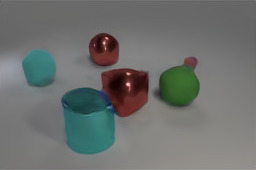} & \includegraphics[width=0.165\linewidth]{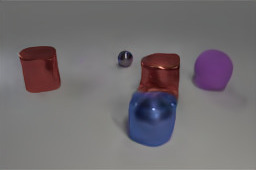} & \includegraphics[width=0.165\linewidth]{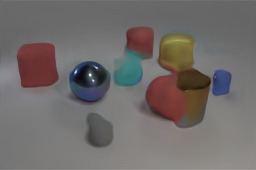} & \includegraphics[width=0.165\linewidth]{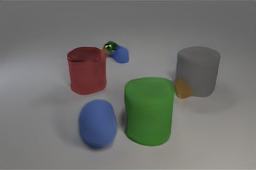} & \includegraphics[width=0.165\linewidth]{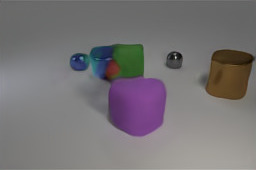} \\

 & \includegraphics[width=0.165\linewidth]{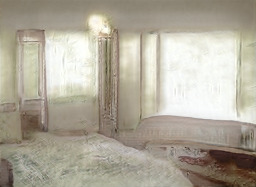} & \includegraphics[width=0.165\linewidth]{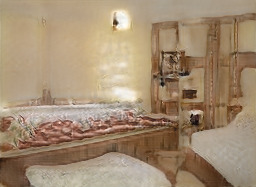} & \includegraphics[width=0.165\linewidth]{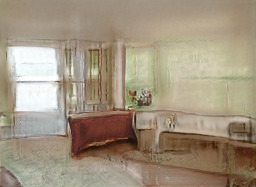} & \includegraphics[width=0.165\linewidth]{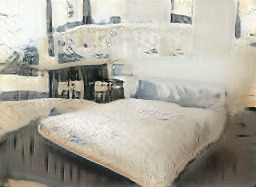} & \includegraphics[width=0.165\linewidth]{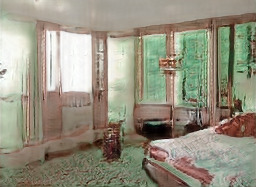} \\

 & \includegraphics[width=0.165\linewidth]{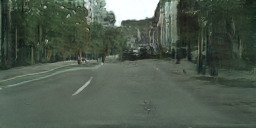} & \includegraphics[width=0.165\linewidth]{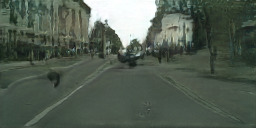} & \includegraphics[width=0.165\linewidth]{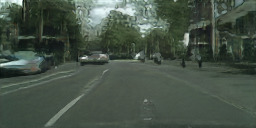} & \includegraphics[width=0.165\linewidth]{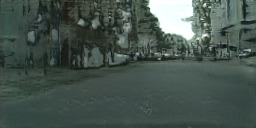} & \includegraphics[width=0.165\linewidth]{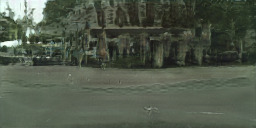} \\

\end{tabular}
\caption{\textbf{State-of-the-art comparison.} A comparison between models' sample images for the CLEVR, LSUN-Bedrooms and Cityscapes datasets. All models have been trained for the same number of steps, which ranges between 5k to 15k kimg training samples. Note that the original StyleGAN2 model has been trained by its authors for up to 70k kimg samples, which is expected to take over 90 GPU-days for a single model. See next pages for comparison with further models. These images show that given the same training length the GANformer model's sample images enjoy higher quality and diversity compared to prior works, demonstrating the efficacy of our approach.}
\label{sota}
\end{figure*}

\begin{figure*}[t]
\centering
\setlength{\tabcolsep}{0pt} 
\renewcommand{\arraystretch}{0} 
\begin{tabular}{c c c c c c}
\rowcolor{white}

\textbf{{SAGAN  }} & \includegraphics[width=0.165\linewidth]{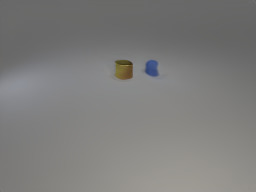} & \includegraphics[width=0.165\linewidth]{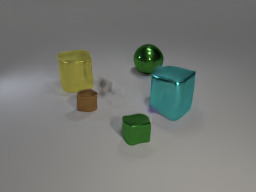} & \includegraphics[width=0.165\linewidth]{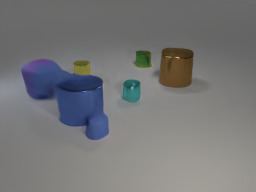} & \includegraphics[width=0.165\linewidth]{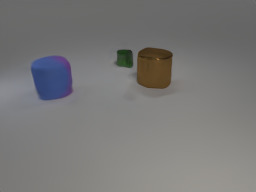} & \includegraphics[width=0.165\linewidth]{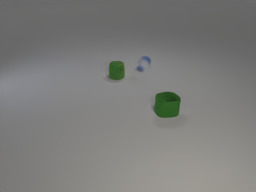} \\

& \includegraphics[width=0.165\linewidth]{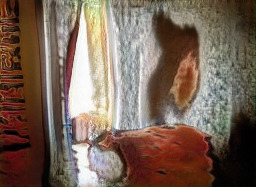} & \includegraphics[width=0.165\linewidth]{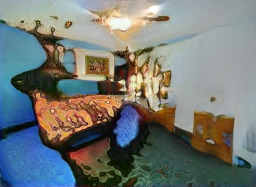} & \includegraphics[width=0.165\linewidth]{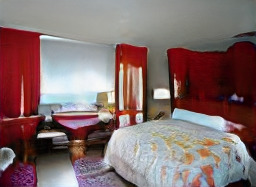} & \includegraphics[width=0.165\linewidth]{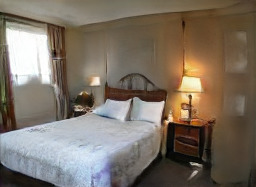} & \includegraphics[width=0.165\linewidth]{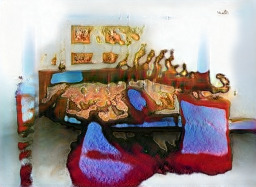} \\

\vspace*{12pt}

 & \includegraphics[width=0.165\linewidth]{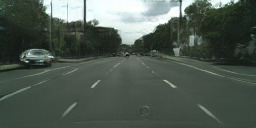} & \includegraphics[width=0.165\linewidth]{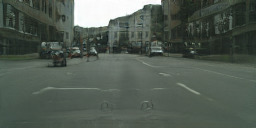} & \includegraphics[width=0.165\linewidth]{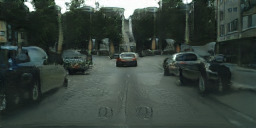} & \includegraphics[width=0.165\linewidth]{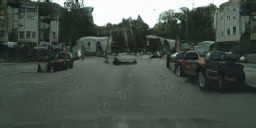} & \includegraphics[width=0.165\linewidth]{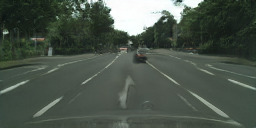} \\

\textbf{{VQGAN  }} & \includegraphics[width=0.165\linewidth]{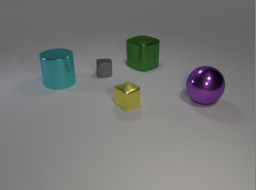} & \includegraphics[width=0.165\linewidth]{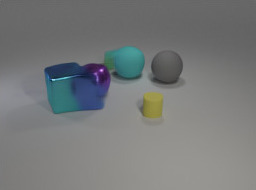} & \includegraphics[width=0.165\linewidth]{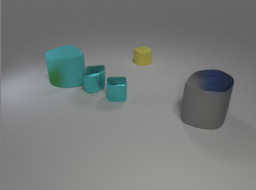} & \includegraphics[width=0.165\linewidth]{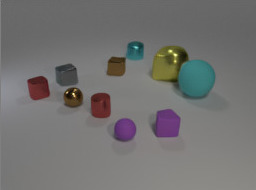} & \includegraphics[width=0.165\linewidth]{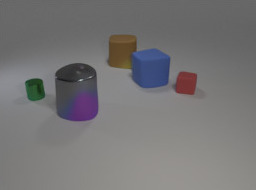} \\

 & \includegraphics[width=0.165\linewidth]{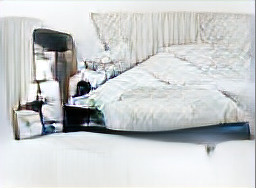} & \includegraphics[width=0.165\linewidth]{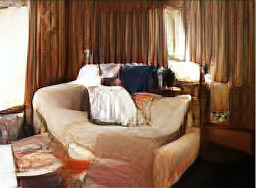} & \includegraphics[width=0.165\linewidth]{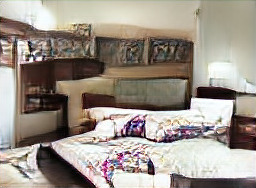} & \includegraphics[width=0.165\linewidth]{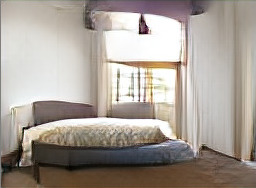} & \includegraphics[width=0.165\linewidth]{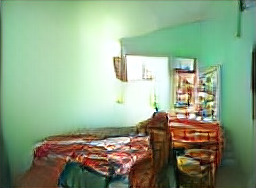} \\

\vspace*{12pt}

 & \includegraphics[width=0.165\linewidth]{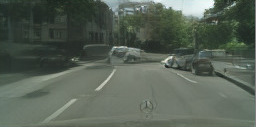} & \includegraphics[width=0.165\linewidth]{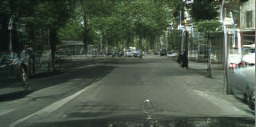} & \includegraphics[width=0.165\linewidth]{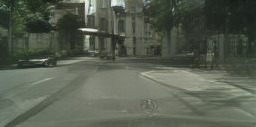} & \includegraphics[width=0.165\linewidth]{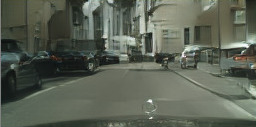} & \includegraphics[width=0.165\linewidth]{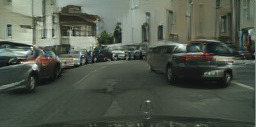} \\

\textbf{{GANformer$_s$  }} & \includegraphics[width=0.165\linewidth]{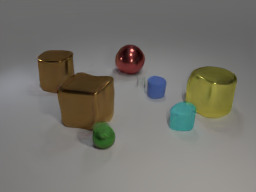} & \includegraphics[width=0.165\linewidth]{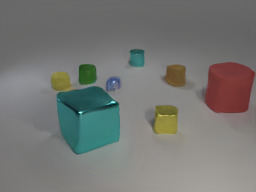} & \includegraphics[width=0.165\linewidth]{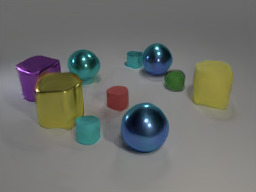} & \includegraphics[width=0.165\linewidth]{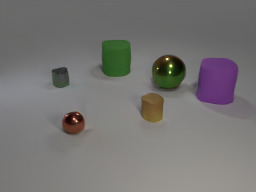} & \includegraphics[width=0.165\linewidth]{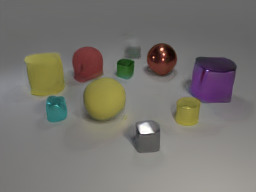} \\

 & \includegraphics[width=0.165\linewidth]{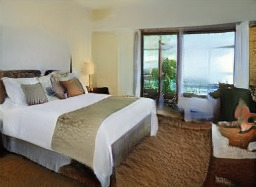} & \includegraphics[width=0.165\linewidth]{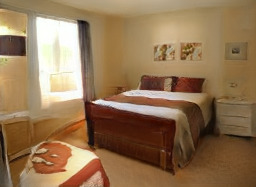} & \includegraphics[width=0.165\linewidth]{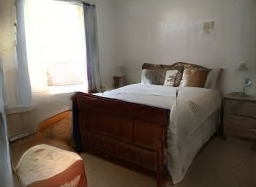} & \includegraphics[width=0.165\linewidth]{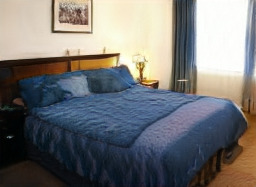} & \includegraphics[width=0.165\linewidth]{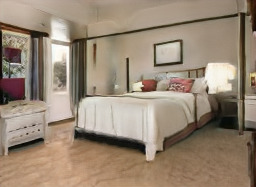} \\

 & \includegraphics[width=0.165\linewidth]{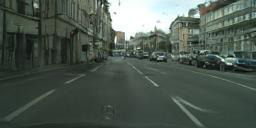} & \includegraphics[width=0.165\linewidth]{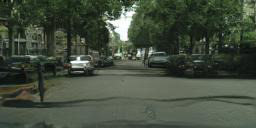} & \includegraphics[width=0.165\linewidth]{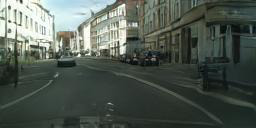} & \includegraphics[width=0.165\linewidth]{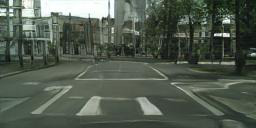} & \includegraphics[width=0.165\linewidth]{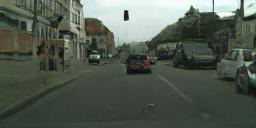} \\

\end{tabular}
\caption{A comparison of models' sample images for the CLEVR, LSUN-Bedrooms and Cityscapes datasets. See figure \ref{sota} for further description.}
\end{figure*}

\begin{figure*}[t]
\centering
\setlength{\tabcolsep}{0pt} 
\renewcommand{\arraystretch}{0} 
\begin{tabular}{c c c c c c}
\rowcolor{white}

\textbf{{GANformer$_d$  }} & \includegraphics[width=0.165\linewidth]{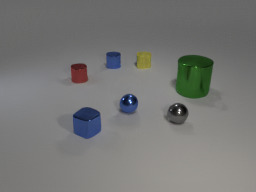} & \includegraphics[width=0.165\linewidth]{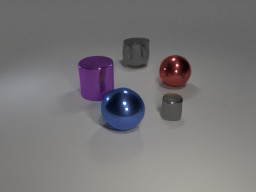} & \includegraphics[width=0.165\linewidth]{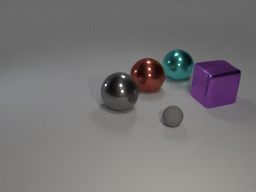} & \includegraphics[width=0.165\linewidth]{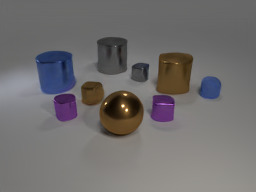} & \includegraphics[width=0.165\linewidth]{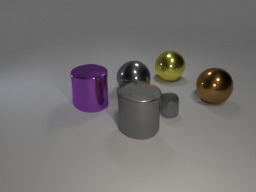} \\

  & \includegraphics[width=0.165\linewidth]{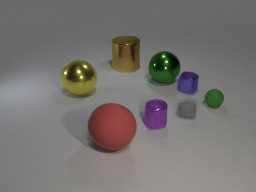} & \includegraphics[width=0.165\linewidth]{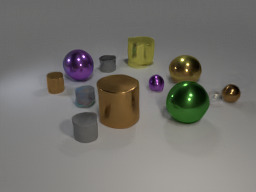} & \includegraphics[width=0.165\linewidth]{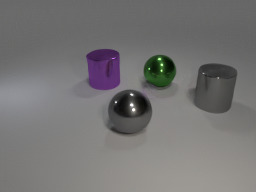} & \includegraphics[width=0.165\linewidth]{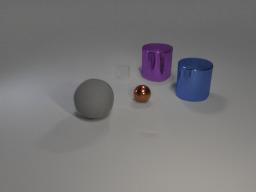} & \includegraphics[width=0.165\linewidth]{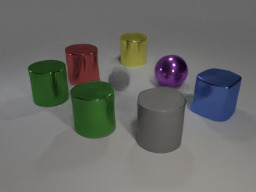} \\

 & \includegraphics[width=0.165\linewidth]{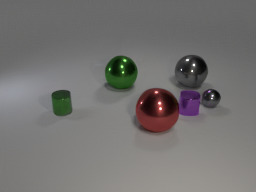} & \includegraphics[width=0.165\linewidth]{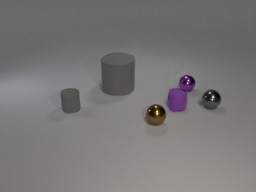} & \includegraphics[width=0.165\linewidth]{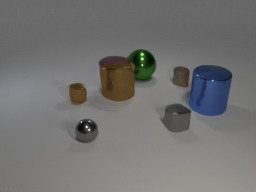} & \includegraphics[width=0.165\linewidth]{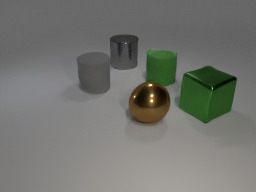} & \includegraphics[width=0.165\linewidth]{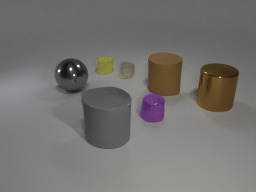} \\
\vspace*{8pt}

 & \includegraphics[width=0.165\linewidth]{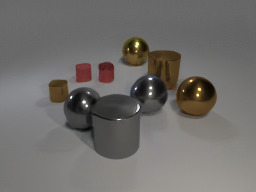} & \includegraphics[width=0.165\linewidth]{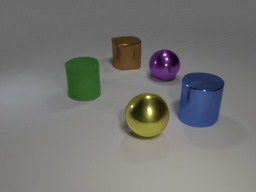} & \includegraphics[width=0.165\linewidth]{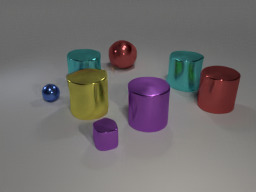} & \includegraphics[width=0.165\linewidth]{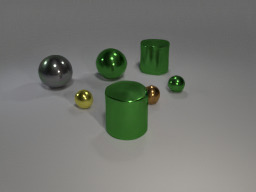} & \includegraphics[width=0.165\linewidth]{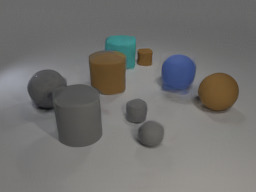} \\

 & \includegraphics[width=0.165\linewidth]{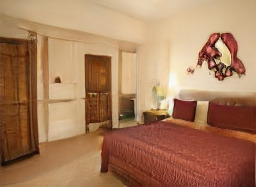} & \includegraphics[width=0.165\linewidth]{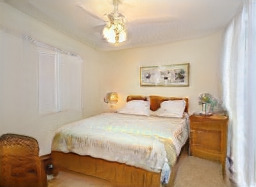} & \includegraphics[width=0.165\linewidth]{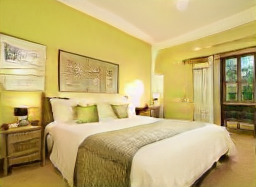} & \includegraphics[width=0.165\linewidth]{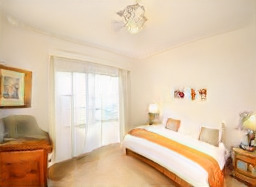} & \includegraphics[width=0.165\linewidth]{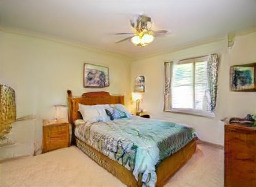} \\

 & \includegraphics[width=0.165\linewidth]{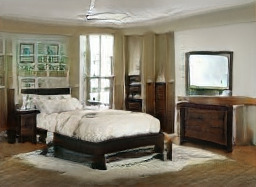} & \includegraphics[width=0.165\linewidth]{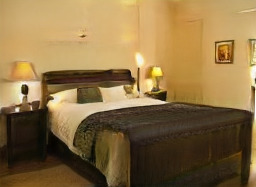} & \includegraphics[width=0.165\linewidth]{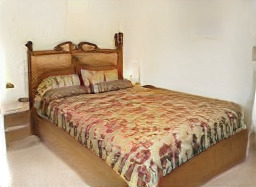} & \includegraphics[width=0.165\linewidth]{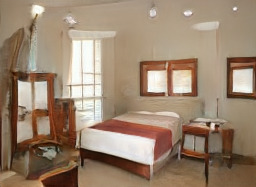} & \includegraphics[width=0.165\linewidth]{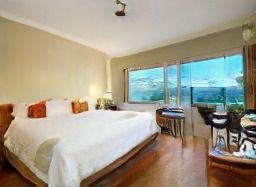} \\

 & \includegraphics[width=0.165\linewidth]{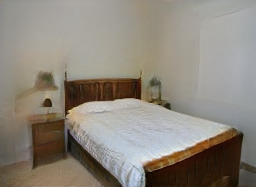} & \includegraphics[width=0.165\linewidth]{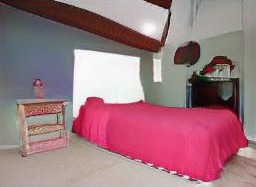} & \includegraphics[width=0.165\linewidth]{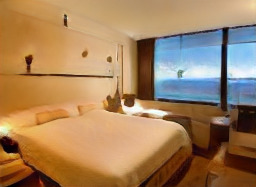} & \includegraphics[width=0.165\linewidth]{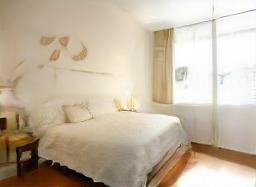} & \includegraphics[width=0.165\linewidth]{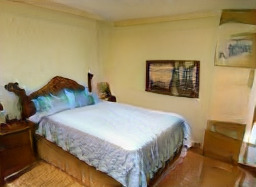} \\
\vspace*{8pt}

 & \includegraphics[width=0.165\linewidth]{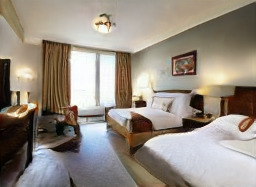} & \includegraphics[width=0.165\linewidth]{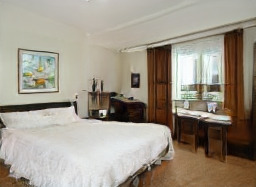} & \includegraphics[width=0.165\linewidth]{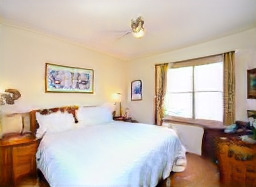} & \includegraphics[width=0.165\linewidth]{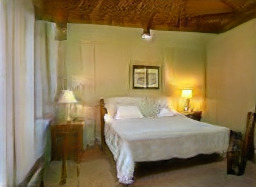} & \includegraphics[width=0.165\linewidth]{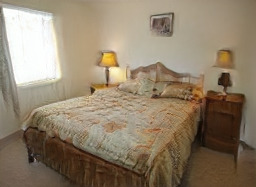} \\

 & \includegraphics[width=0.165\linewidth]{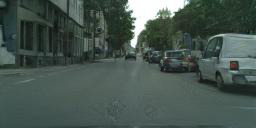} & \includegraphics[width=0.165\linewidth]{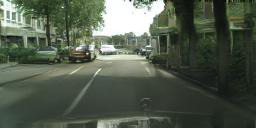} & \includegraphics[width=0.165\linewidth]{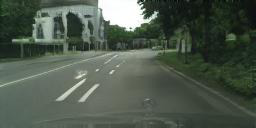} & \includegraphics[width=0.165\linewidth]{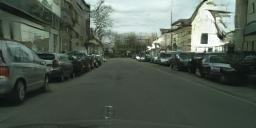} & \includegraphics[width=0.165\linewidth]{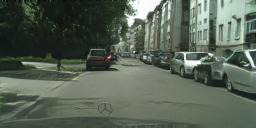} \\

\end{tabular}
\caption{A comparison between models' sample images for the CLEVR, LSUN-Bedrooms and Cityscapes datasets. See figure \ref{sota} for further description.}
\end{figure*}

\begin{figure*}[t]
\centering
\setlength{\tabcolsep}{0pt} 
\renewcommand{\arraystretch}{0} 
\begin{tabular}{c c c c c c}
\rowcolor{white}

\textbf{{GAN  }} & \includegraphics[width=0.165\linewidth]{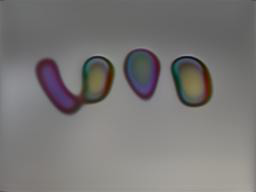} & \includegraphics[width=0.165\linewidth]{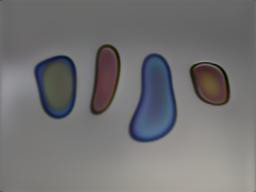} & \includegraphics[width=0.165\linewidth]{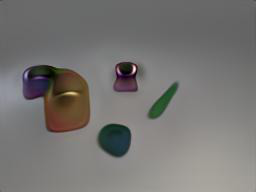} & \includegraphics[width=0.165\linewidth]{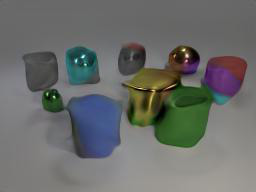} & \includegraphics[width=0.165\linewidth]{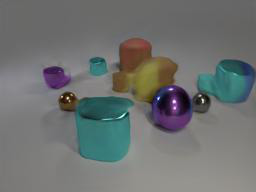} \\

 & \includegraphics[width=0.165\linewidth]{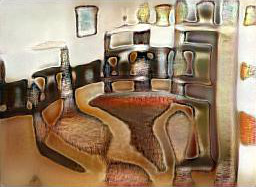} & \includegraphics[width=0.165\linewidth]{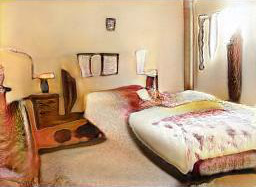} & \includegraphics[width=0.165\linewidth]{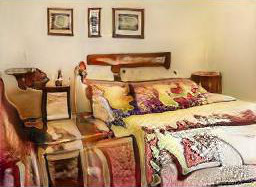} & \includegraphics[width=0.165\linewidth]{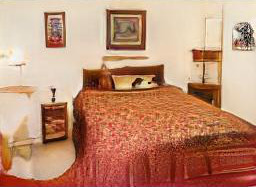} & \includegraphics[width=0.165\linewidth]{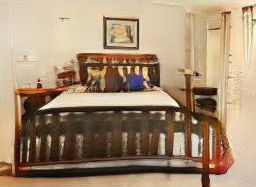} \\

\vspace*{12pt}

 & \includegraphics[width=0.165\linewidth]{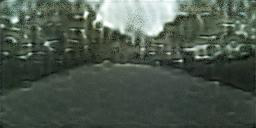} & \includegraphics[width=0.165\linewidth]{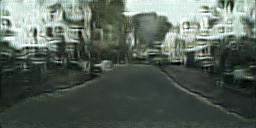} & \includegraphics[width=0.165\linewidth]{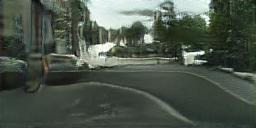} & \includegraphics[width=0.165\linewidth]{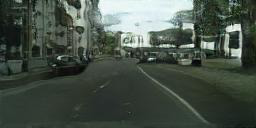} & \includegraphics[width=0.165\linewidth]{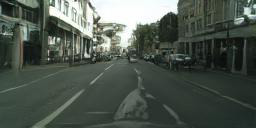} \\

\textbf{{StyleGAN  }} & \includegraphics[width=0.165\linewidth]{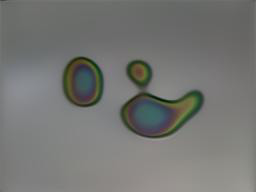} & \includegraphics[width=0.165\linewidth]{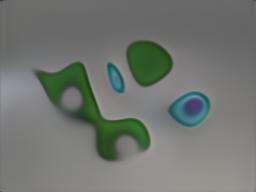} & \includegraphics[width=0.165\linewidth]{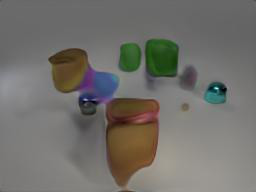} & \includegraphics[width=0.165\linewidth]{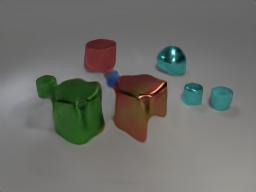} & \includegraphics[width=0.165\linewidth]{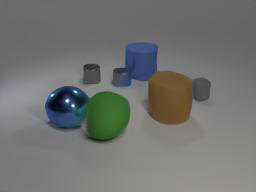} \\

 & \includegraphics[width=0.165\linewidth]{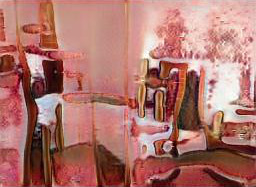} & \includegraphics[width=0.165\linewidth]{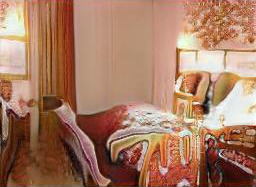} & \includegraphics[width=0.165\linewidth]{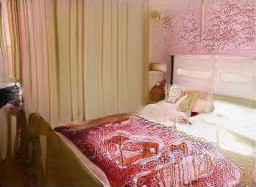} & \includegraphics[width=0.165\linewidth]{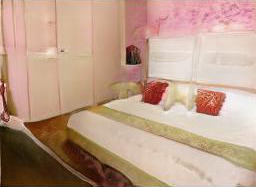} & \includegraphics[width=0.165\linewidth]{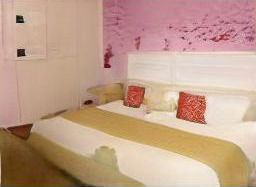} \\

\vspace*{12pt}

 & \includegraphics[width=0.165\linewidth]{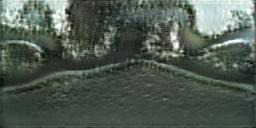} & \includegraphics[width=0.165\linewidth]{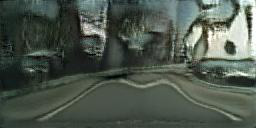} & \includegraphics[width=0.165\linewidth]{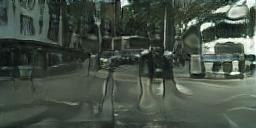} & \includegraphics[width=0.165\linewidth]{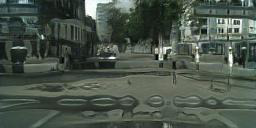} & \includegraphics[width=0.165\linewidth]{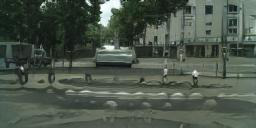} \\

\textbf{{k-GAN  }} & \includegraphics[width=0.165\linewidth]{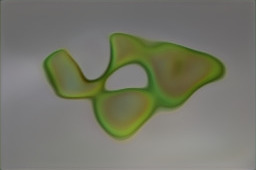} & \includegraphics[width=0.165\linewidth]{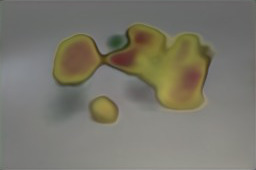} & \includegraphics[width=0.165\linewidth]{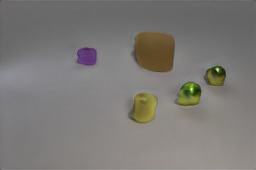} & \includegraphics[width=0.165\linewidth]{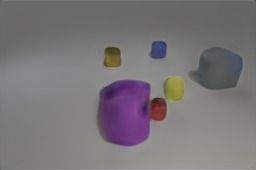} & \includegraphics[width=0.165\linewidth]{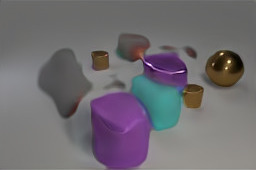} \\

 & \includegraphics[width=0.165\linewidth]{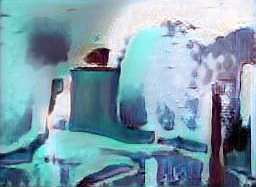} & \includegraphics[width=0.165\linewidth]{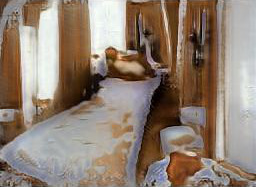} & \includegraphics[width=0.165\linewidth]{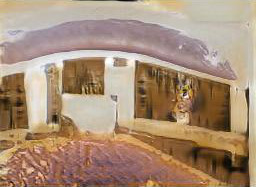} & \includegraphics[width=0.165\linewidth]{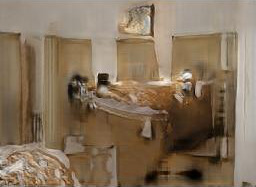} & \includegraphics[width=0.165\linewidth]{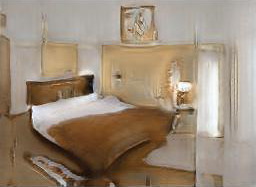} \\

 & \includegraphics[width=0.165\linewidth]{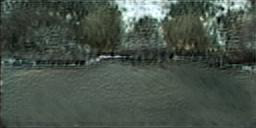} & \includegraphics[width=0.165\linewidth]{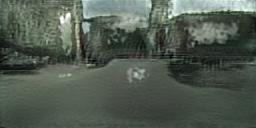} & \includegraphics[width=0.165\linewidth]{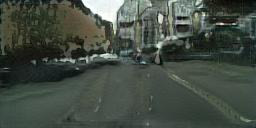} & \includegraphics[width=0.165\linewidth]{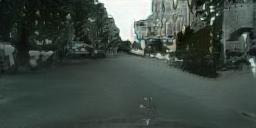} & \includegraphics[width=0.165\linewidth]{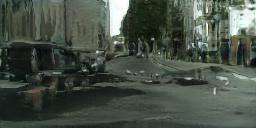} \\

\end{tabular}
\caption{\textbf{State-of-the-art comparison over training.} A comparison between models' sample images for the CLEVR, LSUN-Bedrooms and Cityscapes datasets, generated at different stages throughout the training. Sample images from different points in training are based on the same sampled latent vectors, thereby showing how the image evolves during the training. For CLEVR and Cityscapes, we present results after training to generate 100k, 200k, 500k, 1m, and 2m samples. For the Bedroom case, we present results after 500k, 1m, 2m, 5m and 10m generated samples during training. These results show how the GANformer, especially when using duplex attention, manages to learn a lot faster than competing approaches, generating impressive images early in the training.}
\label{training}
\end{figure*}

\begin{figure*}[t]
\centering
\setlength{\tabcolsep}{0pt} 
\renewcommand{\arraystretch}{0} 
\begin{tabular}{c c c c c c}
\rowcolor{white}

\textbf{{SAGAN  }} & \includegraphics[width=0.165\linewidth]{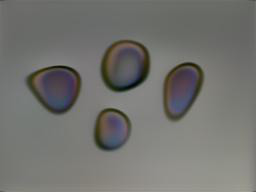} & \includegraphics[width=0.165\linewidth]{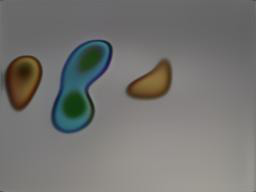} & \includegraphics[width=0.165\linewidth]{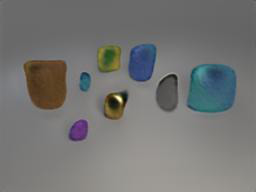} & \includegraphics[width=0.165\linewidth]{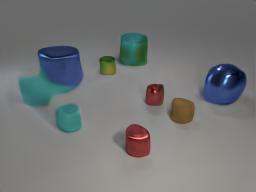} & \includegraphics[width=0.165\linewidth]{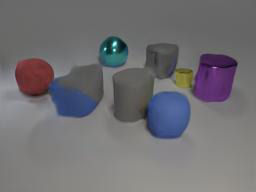} \\

& \includegraphics[width=0.165\linewidth]{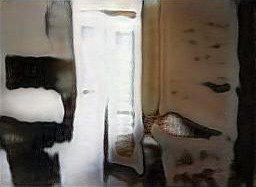} & \includegraphics[width=0.165\linewidth]{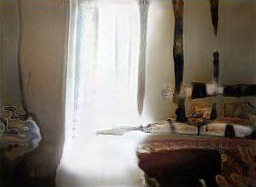} & \includegraphics[width=0.165\linewidth]{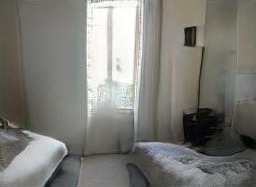} & \includegraphics[width=0.165\linewidth]{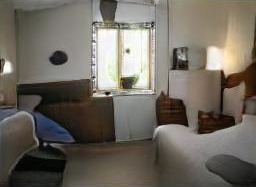} & \includegraphics[width=0.165\linewidth]{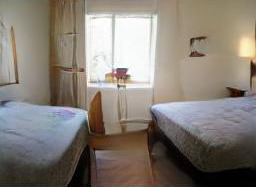} \\

\vspace*{12pt}

 & \includegraphics[width=0.165\linewidth]{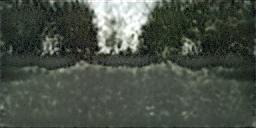} & \includegraphics[width=0.165\linewidth]{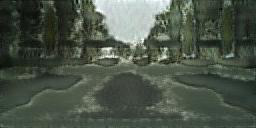} & \includegraphics[width=0.165\linewidth]{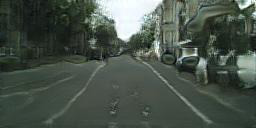} & \includegraphics[width=0.165\linewidth]{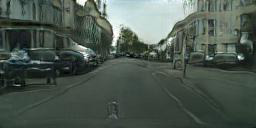} & \includegraphics[width=0.165\linewidth]{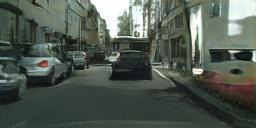} \\

\textbf{{VQGAN  }} & \includegraphics[width=0.165\linewidth]{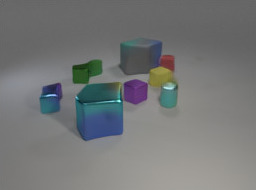} & \includegraphics[width=0.165\linewidth]{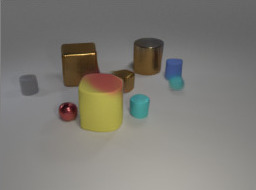} & \includegraphics[width=0.165\linewidth]{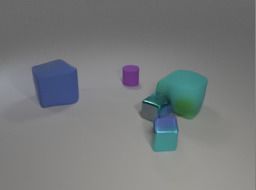} & \includegraphics[width=0.165\linewidth]{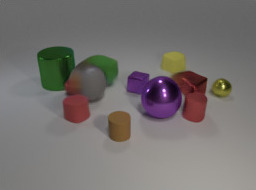} & \includegraphics[width=0.165\linewidth]{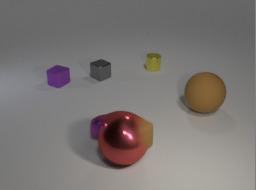} \\

 & \includegraphics[width=0.165\linewidth]{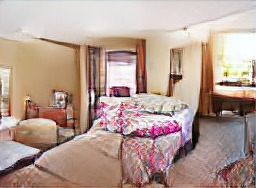} & \includegraphics[width=0.165\linewidth]{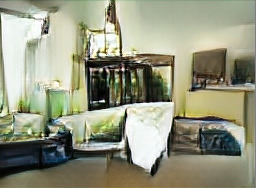} & \includegraphics[width=0.165\linewidth]{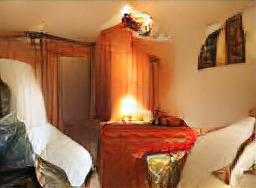} & \includegraphics[width=0.165\linewidth]{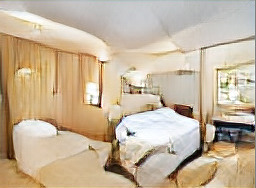} & \includegraphics[width=0.165\linewidth]{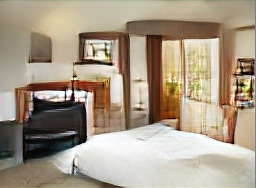} \\

\vspace*{12pt}

 & \includegraphics[width=0.165\linewidth]{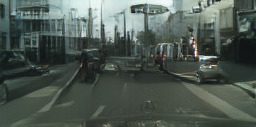} & \includegraphics[width=0.165\linewidth]{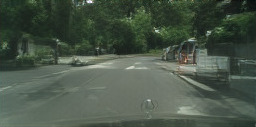} & \includegraphics[width=0.165\linewidth]{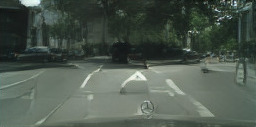} & \includegraphics[width=0.165\linewidth]{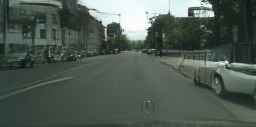} & \includegraphics[width=0.165\linewidth]{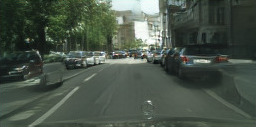} \\

\textbf{{GANformer$_s$  }} & \includegraphics[width=0.165\linewidth]{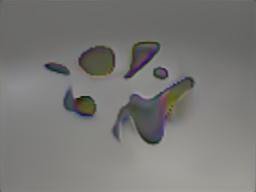} & \includegraphics[width=0.165\linewidth]{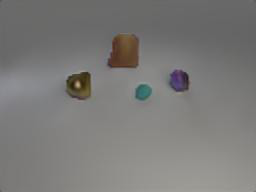} & \includegraphics[width=0.165\linewidth]{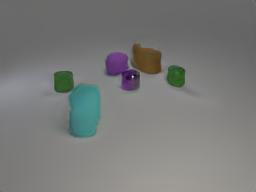} & \includegraphics[width=0.165\linewidth]{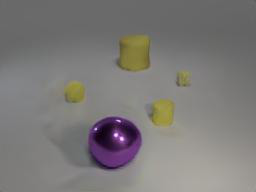} & \includegraphics[width=0.165\linewidth]{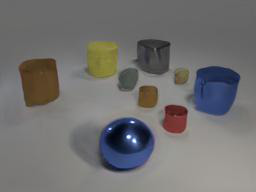} \\

 & \includegraphics[width=0.165\linewidth]{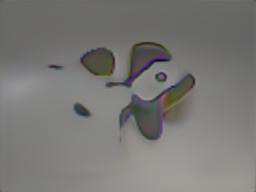} & \includegraphics[width=0.165\linewidth]{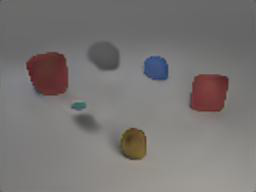} & \includegraphics[width=0.165\linewidth]{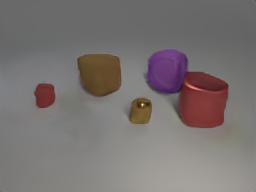} & \includegraphics[width=0.165\linewidth]{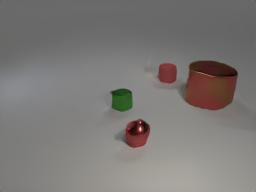} & \includegraphics[width=0.165\linewidth]{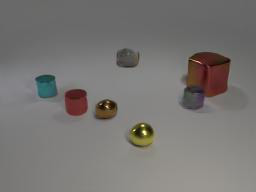} \\

 & \includegraphics[width=0.165\linewidth]{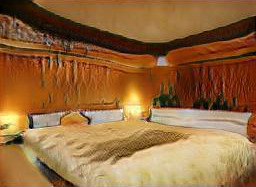} & \includegraphics[width=0.165\linewidth]{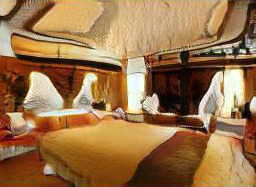} & \includegraphics[width=0.165\linewidth]{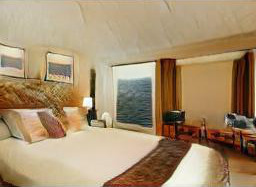} & \includegraphics[width=0.165\linewidth]{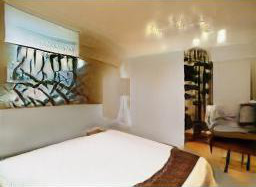} & \includegraphics[width=0.165\linewidth]{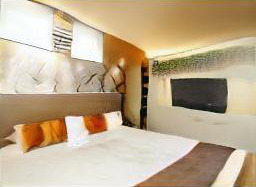} \\

 & \includegraphics[width=0.165\linewidth]{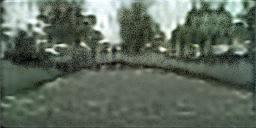} & \includegraphics[width=0.165\linewidth]{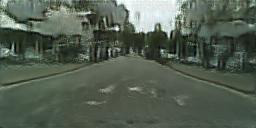} & \includegraphics[width=0.165\linewidth]{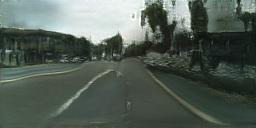} & \includegraphics[width=0.165\linewidth]{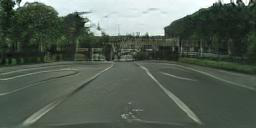} & \includegraphics[width=0.165\linewidth]{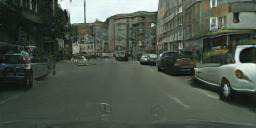} \\

 & \includegraphics[width=0.165\linewidth]{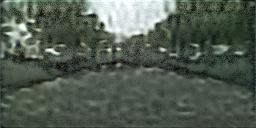} & \includegraphics[width=0.165\linewidth]{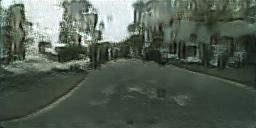} & \includegraphics[width=0.165\linewidth]{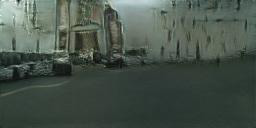} & \includegraphics[width=0.165\linewidth]{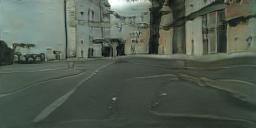} & \includegraphics[width=0.165\linewidth]{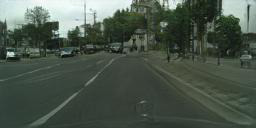} \\

\end{tabular}
\caption{A comparison of models' sample images for the CLEVR, LSUN-Bedrooms and Cityscapes datasets throughout the training. See figure \ref{training} for further description.}
\end{figure*}

\begin{figure*}[t]
\centering
\setlength{\tabcolsep}{0pt} 
\renewcommand{\arraystretch}{0} 
\begin{tabular}{c c c c c c}
\rowcolor{white}

\textbf{{GANformer$_d$  }} & \includegraphics[width=0.165\linewidth]{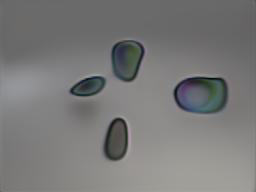} & \includegraphics[width=0.165\linewidth]{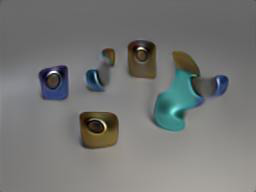} & \includegraphics[width=0.165\linewidth]{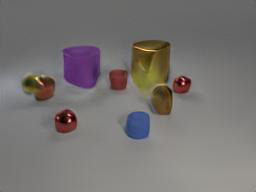} & \includegraphics[width=0.165\linewidth]{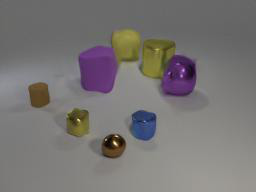} & \includegraphics[width=0.165\linewidth]{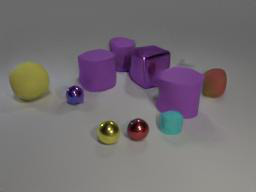} \\
\vspace*{8pt}

  & \includegraphics[width=0.165\linewidth]{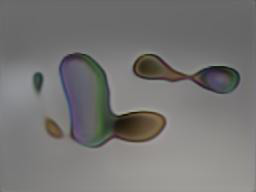} & \includegraphics[width=0.165\linewidth]{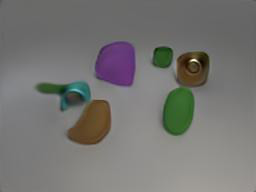} & \includegraphics[width=0.165\linewidth]{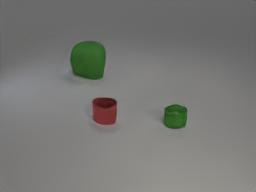} & \includegraphics[width=0.165\linewidth]{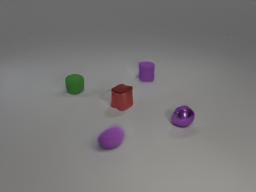} & \includegraphics[width=0.165\linewidth]{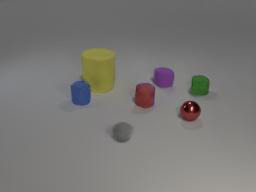} \\

 & \includegraphics[width=0.165\linewidth]{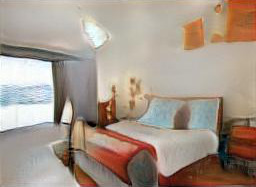} & \includegraphics[width=0.165\linewidth]{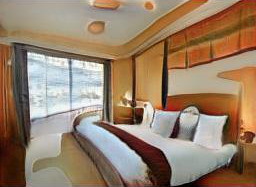} & \includegraphics[width=0.165\linewidth]{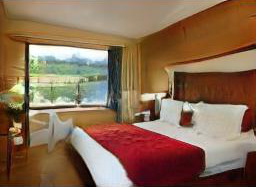} & \includegraphics[width=0.165\linewidth]{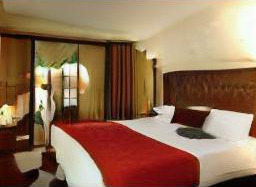} & \includegraphics[width=0.165\linewidth]{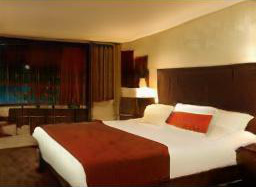} \\
 \vspace*{8pt}
 
 & \includegraphics[width=0.165\linewidth]{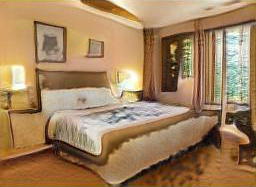} & \includegraphics[width=0.165\linewidth]{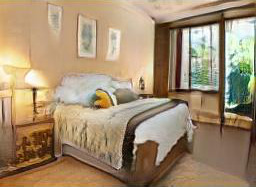} & \includegraphics[width=0.165\linewidth]{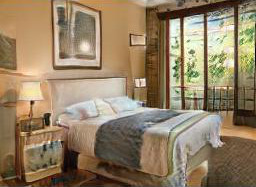} & \includegraphics[width=0.165\linewidth]{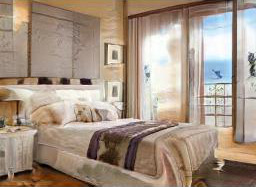} & \includegraphics[width=0.165\linewidth]{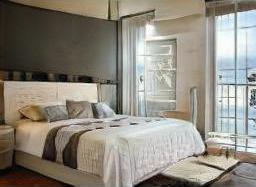} \\

 & \includegraphics[width=0.165\linewidth]{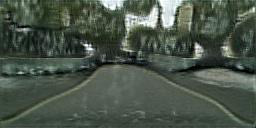} & \includegraphics[width=0.165\linewidth]{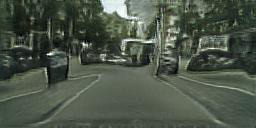} & \includegraphics[width=0.165\linewidth]{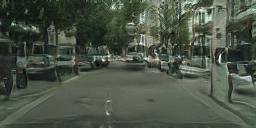} & \includegraphics[width=0.165\linewidth]{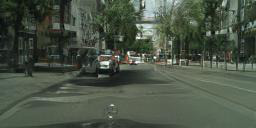} & \includegraphics[width=0.165\linewidth]{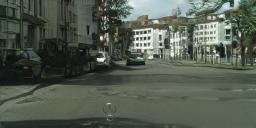} \\

 & \includegraphics[width=0.165\linewidth]{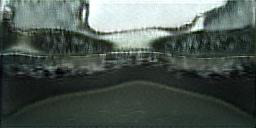} & \includegraphics[width=0.165\linewidth]{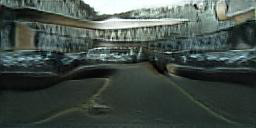} & \includegraphics[width=0.165\linewidth]{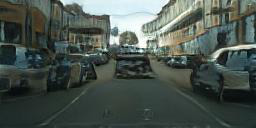} & \includegraphics[width=0.165\linewidth]{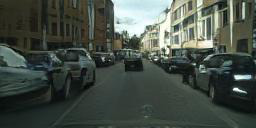} & \includegraphics[width=0.165\linewidth]{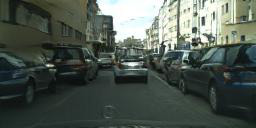} \\

\end{tabular}
\caption{A comparison of models' sample images for the CLEVR, LSUN-Bedrooms and Cityscapes datasets throughout the training. See figure \ref{training} for further description.}
\end{figure*}








\end{document}